\documentclass[journal,twoside,web]{IEEEtran}
\usepackage{cite}
\usepackage{amsmath,amssymb,amsfonts}
\usepackage{algorithmic}
\usepackage{graphicx}
\usepackage{textcomp}
\usepackage{multirow}
\usepackage{pifont}
\usepackage{xcolor, url}

\def\BibTeX{{\rm B\kern-.05em{\sc i\kern-.025em b}\kern-.08em
    T\kern-.1667em\lower.7ex\hbox{E}\kern-.125emX}}

\begin{document}

\title{Resolving the Ambiguity of Complete-to-Partial Point Cloud Registration for Image-Guided Liver Surgery with Patches-to-Partial Matching}

\author{Zixin Yang, Jon S. Heiselman, Cheng Han, Kelly Merrell,  Richard Simon, Cristian. A. Linte.
    \thanks{This work was supported by the National Institutes of Health - National Institute of General Medical Sciences under Award No. R35GM128877 and the National Science Foundation - Division of Chemical, Bioengineering and Transport Systems under Award No. 2245152. }
    \thanks{Zixin Yang, Richard Simon, Kelly Merrell, and Cristian. A. Linte are with the Center for Imaging Science and Department of Biomedical Engineering, Rochester Institute of Technology, Rochester, USA. 
    }
     \thanks{Jon S. Heiselman is with the Department of Biomedical Engineering, Vanderbilt University, USA.
    }
      \thanks{Cheng Han is with the School of Science and Engineering, University of Missouri -- Kansas City, USA.
    }
    \thanks{Email:{ Zixin Yang, 
 yy8898@rit.edu}.}
    } 
    % \thanks{Corresponding author: Zixin Yang.}
   
    \maketitle

\maketitle

% providing the surgeon with subsurface information from preoperative CT/MRI scans during the procedure. 

\begin{abstract}
In image-guided liver surgery, the initial rigid alignment between preoperative and intraoperative data, often represented as point clouds, is crucial for providing sub-surface information from preoperative CT/MRI images to the surgeon during the procedure. Currently, this alignment is typically performed using semi-automatic methods, which, while effective to some extent, are prone to errors that demand manual correction. \textcolor{black}{Alternatively, correspondence-based point cloud registration methods further offer a promising fully automatic solution.} However, they may struggle in scenarios with limited intraoperative surface visibility, a common challenge in liver surgery, particularly in laparoscopic procedures, which we refer to as complete-to-partial ambiguity. We first illustrate this ambiguity by evaluating the performance of state-of-the-art learning-based point cloud registration methods on our carefully constructed \textit{in silico} and \textit{in vitro} datasets. Then, we propose a patches-to-partial matching strategy as a plug-and-play module to resolve the ambiguity, which can be seamlessly integrated into learning-based registration methods without disrupting their end-to-end structure. This approach effectively improves registration performance, especially in low-visibility conditions, \textcolor{black}{reducing registration errors to 6.7 mm (-29\%) \textit{in silico} and 12.5 mm (-40\%) \textit{in vitro}, compared to state-of-the-art performance achieved by Lepard of 9.5 mm and 20.7 mm, respectively.} The constructed benchmark and the proposed module establish a solid foundation for advancing applications of point cloud correspondence-based registration methods in image-guided liver surgery. Our code and datasets will be released at \url{https://github.com/zixinyang9109/P2P}.

% For example, under low organ visibility, the approach improves registration errors from LiverMatch from 12.9 to 8.7 mm (-33\%)/20.5 to 15.0 (-27\%) and Lepard from 9.5 to 6.7 mm (-29\%)/20.7 to 12.5 (-40\%) on in silico/in vitro datasets, respectively, with better noise robustness.

\end{abstract}

\begin{IEEEkeywords}
Image-guided liver surgery, point cloud matching, pre- to intraoperative rigid registration.
\end{IEEEkeywords}

\section{Introduction}
\label{sec:introduction}

\begin{figure}[!t]
\centering
\includegraphics[width=0.9\linewidth]{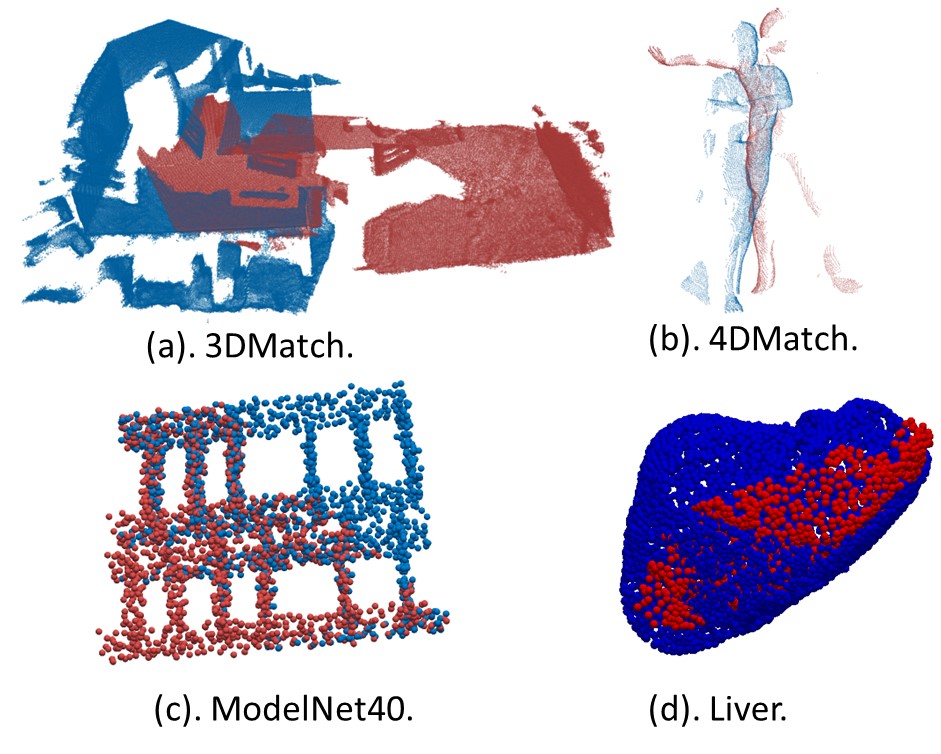}
\caption{Comparison of widely used public point registration datasets in computer vision (3DMatch, 4DMatch, ModelNet40) and our liver registration dataset. In each example, the source point clouds (blue) are aligned with the target point clouds (red). 3DMatch, 4DMatch, and ModelNet40 focus on partial-to-partial cases. The liver registration dataset is distinct from these datasets, designed under the complete-to-partial scenarios.
 }
\label{fig:Datasets_comparison}
\end{figure}

\IEEEPARstart{I}{n} liver surgery, preoperative imaging techniques such as Computed Tomography (CT) and Magnetic Resonance Imaging (MRI) provide detailed information about critical anatomical structures, such as blood vessels and tumors. However, this information is \textcolor{black}{typically unavailable} during surgery due to the large footprint of the equipment, its immobility, and its prohibitively high cost. Consequently, intraoperative imaging is most commonly limited to lower resolution or partial information covering only a portion of the visible organ from modalities including intraoperative ultrasound \cite{heiselman2020intraoperative, smit2023ultrasound}, cone-beam CT (CBCT) \cite{peterlik2018fast, ivashchenko2021cbct}, and optical cameras \cite{rucker2013mechanics, koo2022automatic, suwelack2014physics, golse2021augmented}.
To inform the surgeon of subsurface information from preoperative CT/MRI during surgery, a registration process is involved to compute the transformation between the preoperative and intraoperative data in image-guided surgery (IGS) systems, which is generally divided into two steps.

% \textcolor{blue}{
% The first step in the registration process is the 
Step 1 is an initial rigid registration to resolve the problem that preoperative and intraoperative data are acquired in different coordinate systems. However, current IGS systems often perform this alignment semi-automatically \cite{acidi2023augmented,teatini2019effect}, bringing considerable time consumption and susceptibility to errors \cite{teatini2019effect}.
% Current methods,Current IGS systems, however, often perform this alignment manually \cite{acidi2023augmented}, bringing
% with considerable time consumption and susceptibility to errors \cite{teatini2019effect}. 
% Different from current IGS systems, the preoperative and intraoperative data are acquired in different coordinate systems for liver surgery.
% However, current IGS systems often perform this alignment Semi-automatically \cite{acidi2023augmented}, bringing considerable time consumption and susceptibility to errors \cite{teatini2019effect}. 

Step 2 involves non-rigid registration methods to estimate organ deformations, which depend on an accurate initial rigid registration.

Registration methods that adapt to available intraoperative information can be categorized into intensity-based, 3D-to-2D, and 3D-to-3D methods. This paper addresses the initial rigid registration problem in a 3D-to-3D context, where both the preoperative and intraoperative data are represented as 3D point clouds. 
The preoperative point cloud is often derived from CT or MRI scans, while the intraoperative point cloud is often captured using 3D sensors or reconstructed from endoscopic video images via 3D reconstruction. For clarity, we refer to the point cloud representing the entire liver from the preoperative surface as the \textbf{source point cloud} and the partial intraoperative point cloud as the \textbf{target point cloud}.

Learning-based point cloud registration methods, particularly correspondence-based approaches, have shown promising results \cite{zeng20173dmatch, huang2021predator, li2022lepard, yu2021cofinet} to solve the limitations of semi-automatic methods. However, applying these methods, designed for natural datasets,  to liver surgery is challenging due to two reasons:
% where the visible intraoperative surface is limited—introduces two significant challenges that hinder achieving the same level of accuracy.

\textbf{I. Insufficient intraoperative-oriented design}. Most learning-based point cloud registration methods are designed for partial-to-partial cases, where point cloud pairs are captured from different view angles and have roughly similar scales. These pairs typically come from RGB-D cameras \cite{huang2021predator} (Fig. \ref{fig:Datasets_comparison} (a)), simulated depth maps from animated sequences \cite{li2022lepard} (Fig. \ref{fig:Datasets_comparison} (b)), or synthetic objects \cite{wu20153d, qi2017pointnet} (Fig. \ref{fig:Datasets_comparison} (c)). Those pairs often feature prominent structures, such as edges and corners from natural scenes.  

In liver surgery, particularly in laparoscopic liver surgery, only a partial liver surface is visible intraoperatively: typically around 20-30$\%$ of the surface \cite{heiselman2018characterization} is visible, \textcolor{black}{as laparoscopic cameras provide a narrow and localized field of view compared to open surgery.} The partially observable liver surface is matched against the complete liver surface segmented from the preoperative images. Since the viewpoint can change during the procedure, i.e., there is no strong prior knowledge as to what sub-region of the preoperative liver surface should match the intraoperative point cloud match. 

As a further challenge, the smooth surface of the liver lacks strong visual key points that could assist with matching partial point clouds and complicates current approaches to accurately align the visible intraoperative portion with the complete preoperative liver model (i.e., see Fig. \ref{fig:Datasets_comparison} (d), where the target point cloud has a limited visible surface area). Besides the confusion associated with matching to the top or bottom of the liver, there is also ambiguity when registering a small point cloud to regions adjacent to the liver dome.

As such, we argue that current learning-based point cloud registration methods are insufficient to address the complete-to-partial ambiguity that frequently arises in \textcolor{black}{laparoscopic liver surgery}. \textcolor{black}{
Traditional ICP-based methods \cite{besl1992method,yang2015go}, often regarded as alternative solutions, also face limitations, including the need for manual intervention and high computational cost.
% Traditional ICP-based methods \cite{besl1992method,yang2015go}, usually considers as alternatives, also have limitations, requiring manual interaction and being computationally expensive. 
Meanwhile, handcrafted features \cite{dos2014pose} struggle with the challenges posed by smooth surfaces that lack strong edges or curvature and lead to inherent ambiguity in point cloud matching.}

\textbf{II. Substantial data requirements}. 
The success of learning-based point cloud registration methods is heavily based on scaled-up datasets. However, this 
is particularly challenging for liver surgery as collecting extensive \textit{in vivo} or \textit{in vitro} datasets is expensive, complex, and time-consuming. 
A common practice to capture the preoperative and intraoperative configurations of the liver is to embed fiducial markers. However, this process may cause damage to tissues or phantoms.
%, and establishing correspondences of markers is not trivial
% are usually embedded to ; 
% However, this process can damage tissues or phantoms, and identifying the correspondence of markers is not trivial. 
An alternative is to use simulated data for training and then apply the model to real-world \textit{in vivo} or \textit{in vitro} data \cite{pfeiffer2020non}. 
% Although promising, there is currently a lack of large, publicly accessible simulated datasets, and the availability of real-world testing datasets remains severely constrained.
While promising, there is currently a scarcity of large, publicly accessible simulated datasets, and the availability of real-world testing datasets remains severely constraining.

% Though promising, there is currently no large, publicly available simulated dataset, and the availability of real-world testing datasets is extremely limited. 
% Nevertheless, a large public dataset is essential for training, as well as for ensuring fair comparison of different methods, which is crucial for the advancement of learning-based techniques.

\subsection{Contribution}

To address the first challenge, we propose a novel, learnable-parameter-free approach that integrates seamlessly with correspondence-based point cloud registration methods to mitigate complete-to-partial ambiguity (see \S\ref{subsec:patches}). Our approach significantly improves low-visibility performance with negligible computational overhead. The core concept is to transform the complete-to-partial point cloud registration problem into a subdivided matching problem between the partial point cloud and multiple patches extracted from the complete cloud, which we refer to as patches-to-partial (P2P) matching.

\textcolor{black}{In the P2P module, we first identify the most likely visible points from the source point cloud by selecting the top points with the highest visibility scores, computed by summing feature similarities (see \S\ref{subsec:p2p1}). Patch nodes are then generated from these points and subsequently used to create candidate patches, each containing the same number of points as the target point cloud (see \S\ref{subsec:p2p2} and \ref{subsec:p2p3}).} Finally, we match the target point cloud to these candidate patches and aggregate a robust, rigid point cloud transformation by selecting the best estimate across candidates (see \S\ref{subsec:p2p4} and \ref{subsec:p2p5}).This module only involves feature resampling and rematching, allowing it to be parallelized for fast execution. Additionally, it is fully differentiable, making it suitable for integration into different correspondence-based point cloud and learning-based registration pipelines.

To address the second challenge, we developed a large simulated dataset comprising over $1,000$ liver models and $10,000$ simulations alongside phantom datasets (see \S\ref{sec:dataset}), providing valuable resources for benchmarking and future advancing research in the liver registration area. 

We experimentally demonstrate that the current state-of-the-art correspondence-based registration methods are not sufficient when dealing with complete-to-partial ambiguity. 
% and verify their performances on the proposed datasets, illustrating their shortcomings when dealing with complete-to-partial ambiguity. 
We then show the effectiveness of our proposed module in dealing with low-visibility cases after seamless plug-and-play integration
into two promising methods, \textcolor{black}{ LiverMatch\cite{yang2023learning} and Lepard \cite{li2022lepard} (for additional details on the selection of these methods, refer to \S\ref{sec:dicsA})}. Without any adjustment to loss functions, additional learnable modules, outlier rejection methods, etc., our approach significantly improves complete-to-partial registration.

% achieving performance that is comparable to or surpasses that of the commonly utilized robust estimator, RANSAC.

% even on par with or better than those improved with the typically used robust estimator, RANSAC.

% Moreover, as learning-based point cloud registration methods are sensitive to hte
% the proper pre-processing steps that enable point clouds from simulations and real dataset
% large public datasets for training and validating learning-based methods in liver surgery still do exist yet.
% biomechanical model
% For geometry.

% as both
% .
% unlike
% This gap arises from the challenges of 
% gathering a substantial amount of in vivo or in vitro data for training learning-based methods. 

%  Furthermore, there's a noticeable gap in current research concerning learning-based correspondence \cite{yang2023learning} and registration techniques \cite{pfeiffer2020non, labrunie2023automatic, guan2023intraoperative} in the context of liver surgery.

\section{Related work}

\subsection{Traditional point cloud registration}

Iterative Closest Point (ICP) \cite{besl1992method} and its variants are commonly used in semi-automatic registration processes. ICP iteratively finds the closest points between two point clouds and estimates the rigid transformation using singular value decomposition (SVD) \cite{arun1987procrustes}. Though intuitive, ICP is sensitive to initial alignment. 
% To address ICP's limitations, g
Globally optimal ICP (GO-ICP) \cite{yang2015go} poses an alternative approach by introducing a global search strategy to improve robustness. However, 
% while this approach improves robustness, 
it then becomes computationally expensive and does not always guarantee an optimal solution.

\subsection{Learning-based point cloud registration}\label{subsec:learning-based}

Learning-based point cloud registration methods can be broadly categorized into correspondence-based and direct registration approaches. Correspondence-based methods \cite{huang2021predator, li2022lepard} first establish corresponding point pairs between two point clouds, then compute the optimal transformation based on these correspondences. Direct registration methods \cite{aoki2019pointnetlk, xu2021omnet}, on the other hand, directly regress the rigid transformation from global features. While these methods show promising results on synthetic, object-centric point clouds \cite{wu20153d}, they usually struggle with low overlap ratios, limited transformation ranges, and larger-scale scenes \cite{huang2021predator, yew2022regtr}. Currently, the state-of-the-art techniques assessed on publicly available benchmarking datasets \cite{zeng20173dmatch, huang2021predator, li2022lepard} are correspondence-based.

% \textcolor{blue}{
Representative correspondence-based methods, such as 3DMatch \cite{zeng20173dmatch}, Predator \cite{huang2021predator}, and Lepard \cite{li2022lepard}, introduce techniques like Siamese networks, attention modules, and positional encoding to improve point cloud matching. These techniques and their variants have since become standard practice in correspondence-based methods. While most of these methods focus directly on fine-scale correspondences, some, such as those reported in \cite{yu2021cofinet, qin2022geometric, yu2023rotation}, adopt a point-to-node grouping strategy, where coarse-scale matches guide fine-scale correspondences within patches. 
% \textcolor{blue}{
However, whether this strategy performs well on datasets with complete-to-partial ambiguity remains unclear. Thus, a representative method \cite{yu2023rotation} using the strategy is included in the experiments.

% Although this strategy works well when the source and target point clouds are of similar scale, 

% it struggles in low-visibility cases, such as in liver surgery, where sparse coarse-scale point clouds increase the risk of matching errors and outliers at the fine scale.

Early correspondence-based methods~\cite{zeng20173dmatch,huang2021predator} rely on RANSAC, a widely known robust estimator that iteratively selects subsets of correspondences to find the best transformation. However, RANSAC often suffers from slow convergence \cite{zhang20233d} and is unsuitable for end-to-end learning schemes. Consequently, recent methods aim to achieve comparable results without relying on RANSAC. For instance, RegTr~\cite{yew2022regtr} directly predicts keypoint displacements, while RoITr \cite{yu2023rotation} uses a local-to-global registration approach to eliminate the need for RANSAC.

Overall, correspondence-based methods have a strong foundation in several public benchmarks for natural datasets. Still, they are not designed to handle the complete-to-partial cases encountered in liver surgery.

\subsection{3D-3D Initial Rigid Registration in Liver Surgery}

\textcolor{black}{3D-3D initial rigid registration for liver surgery often relies on manual adjustment \cite{acidi2023augmented,teatini2019effect} or landmark-based semi-automatic alignment \cite{feuerstein2008intraoperative, herline2000surface}. However, manual adjustment is prone to user-dependent errors. At the same time, anatomical landmarks can be difficult and ambiguous to annotate due to occlusions from surrounding anatomical structures and the smooth surface of the liver during surgery.}

Automatic correspondence methods using hand-crafted feature descriptors have been explored by Dos Santos \textit{et al.} \cite{dos2014pose} and Robu \textit{et al.} \cite{robu2018global}. However, these methods have limited accuracy and often require additional pruning.

Several learning-based methods \cite{guan2023intraoperative, yang2023learning, zhang2024point} have been proposed for 3D-3D liver registration, yielding promising results. However, these outcomes should be interpreted with caution. For example, the simulation dataset used in \cite{yang2023learning} assumes one-to-one correspondence, as both the source and target point clouds are derived from the same mesh without post-processing. Consequently, models pre-trained or fine-tuned on these datasets may struggle when applied to real-world data. Furthermore, Guan \textit{et al.} \cite{guan2023intraoperative} and Zhang \textit{et al.} \cite{zhang2024point} used the intraoperative CT surface as the preoperative surface in the porcine dataset \cite{modrzejewski2019vivo}, resulting in source and target point clouds with no deformation. 

In contrast, we carefully post-process our simulations to include soft tissue deformations, eliminate one-to-one correspondences, and remove rigid components from the simulated deformations (\textcolor{black}{see \S\ref{subsubsec:in_silico_dataset}}). We evaluate the selected registration methods on both our simulation and phantom datasets, ensuring a more rigorous and realistic assessment.

% with only limited transformations.

\section{Methodology}
\label{sec:method}

\begin{figure*}[!t]
    \centering
\includegraphics[width=0.85\linewidth]{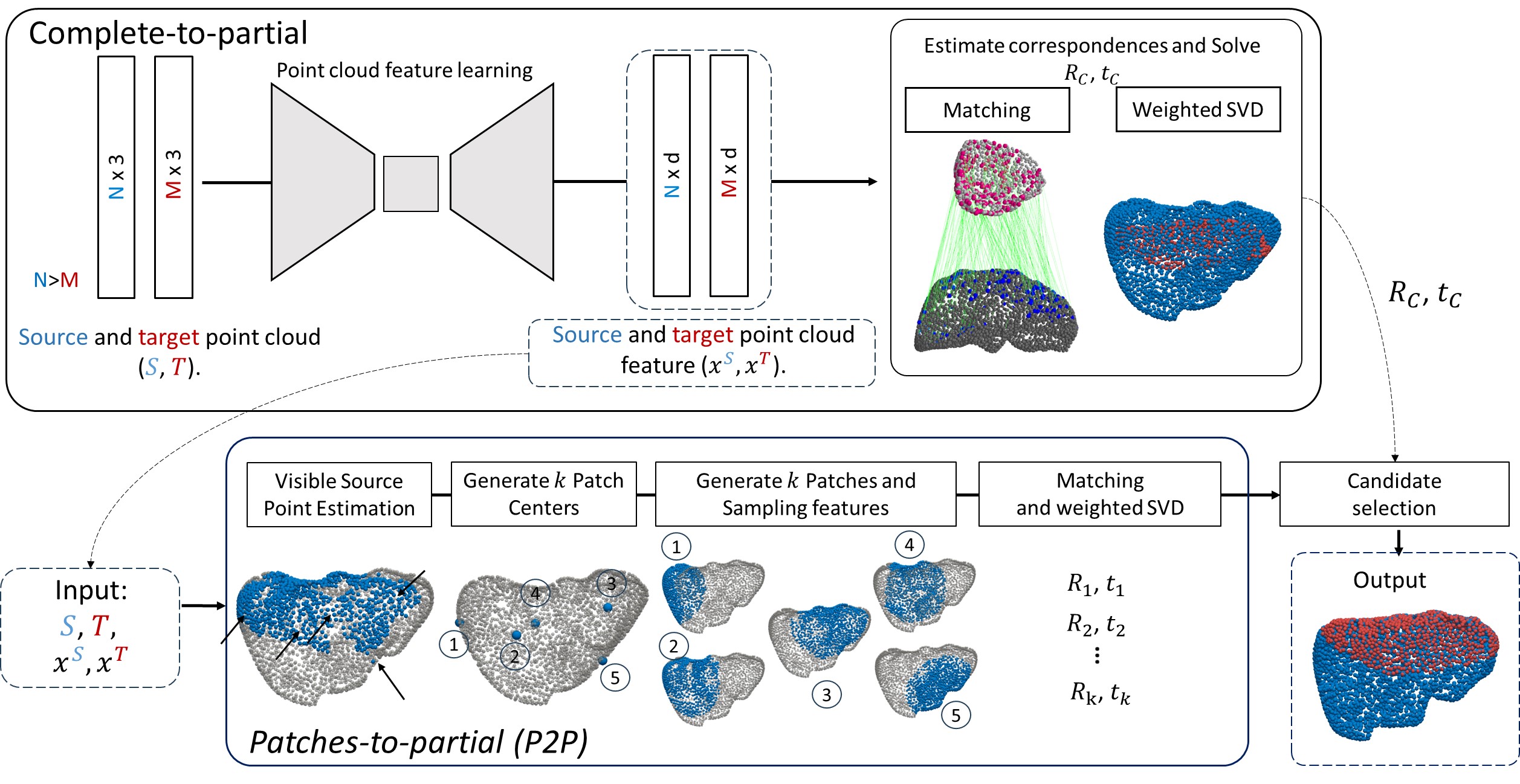}
    \caption{Illustration of the general paradigm of learning-based correspondence registration methods (top) and our plug-and-play P2P module (bottom). \textcolor{black}{The input source and target point clouds are voxelized to maintain similar densities, ensuring a uniform and continuous representation of the liver surface regions (see \S\ref{subsec:Preprocessing} and \S\ref{sec:dicsB}).} Our proposed module generates candidate patches, samples their point-wise features from a learning-based correspondence registration method, and performs feature matching and rigid transformation estimation for each patch with the target point cloud. Finally, a candidate selection rule determines the optimal rigid transformation. We present the case where the patch number $k$ is set to 5. \textcolor{black}{The patch consists of the same number of points as the target point cloud (see \S\ref{subsec:patches} and \S\ref{subsec:p2p3}).}
    }
    \vspace{-1.5em}
    \label{fig:framework}
\end{figure*}

This section introduces our proposed P2P strategy as a plug-and-play module that can be seamlessly integrated into current learning-based correspondence registration methods. We begin by presenting the general paradigm of learning-based correspondence registration methods (see \S\ref{subsec:prelim}), followed by a detailed module explanation (see \S\ref{subsec:patches}). The overall framework is illustrated in    Fig.~\ref{fig:framework}.
  
\subsection{Preliminaries}\label{subsec:prelim}

Given a source point cloud $\mathbf{S} \in \mathbb{R}^{N\times3}$ (preoperative) and a target point cloud $\mathbf{T} \in \mathbb{R}^{M\times3}$ (intraoperative) as inputs, correspondence-based registration methods output their respective point-wise features, $\mathbf{x}^{\mathbf{S}} \in \mathbb{R}^{N \times d}$ and $\mathbf{x}^{\mathbf{T}} \in \mathbb{R}^{M \times d}$, along with the rigid transformation from $\mathbf{S}$ to $\mathbf{T}$. Here, $N$ and $M$ represent the number of participating points in the source and target point clouds, respectively, and $d$ is the feature dimension. 
These registration methods generally consist of four key components: \textit{Feature extraction network}, \textit{Matching module}, \textit{Rigid transformation estimation module}, and \textit{Loss functions.}

\textit{Feature extraction networks} extract point-wise feature descriptors from both the source and target point clouds. These networks typically use a point cloud feature learning backbone, such as PointNet++ \cite{qi2017pointnet++}, KPConv \cite{thomas2019kpconv}, or Point Transformer \cite{zhao2021point}, which usually involves a U-Net-like structure that gradually down-samples the input point clouds to extract features, followed by upsampling to restore the resolution of the original points while preserving the extracted features.

Attention mechanisms, which have emerged as a pivotal component for feature understanding in various domains~\cite{huang2021predator, li2022lepard, yu2023rotation}, are frequently incorporated within the feature learning backbone, particularly in the bottleneck, to empower the interaction between local geometric features towards global geometry-aware ones. This interaction can occur either within the same point cloud (self-attention) or across different point clouds (cross-attention), depending on the layer used.

% For example, given input feature $\mathbf{x}^{\mathbf{S}}$, $\mathbf{x}^{\mathbf{S}}$ is firstly projected to query vector $\mathbf{q}$, key vector $\mathbf{k}$, and  value vector $\mathbf{v}$. The attention operation identifies relevant information by assessing how similar the query vector $\mathbf{q}$ is to the key vector $\mathbf{k}$. The final output vector is then formed by combining the value vector $\mathbf{v}$, each value adjusted based on the similarity scores $a$.

\textit{Matching modules} then use either Sinkhorn optimal transport algorithm \cite{sinkhorn1967concerning, yu2023rotation} or dual-softmax operation \cite{rocco2018neighbourhood,li2022lepard} to match the source and target feature descriptors, which output correspondences between source and target point clouds and their confidences. 

\textit{Rigid transformation estimation modules} further utilize the estimated correspondences and confidences from the matching module and calculate the rigid transformation, using the SVD or weighted SVD \cite{arun1987least,besl1992method}. If the estimated correspondences contain a large portion of outliers, the weighted SVD tends to predict an inaccurate rigid transformation. In this case, additional outlier rejection methods are involved in sampling inlier correspondence (e.g., RANSAC).

\textit{Loss functions} consist of one or more types, such as classification-based loss\cite{li2022lepard} (e.g., cross-entropy with dual-softmax or optimal transport), feature metric-based loss\cite{huang2021predator,yu2023rotation} (e.g., circle loss), and displacement-based loss for predicted matches or estimated rigid transformation accuracy in 3D space.

\subsection{Transferring Complete-to-Partial into P2P Matching}
\label{subsec:patches}

% Though correspondence-based registration is well-established in computer vision (see \S\ref{subsec:prelim}), 
% it fails to address the complete-to-partial ambiguity inherent in liver surgery. The disparities between ideal registration scenarios and \textcolor{black}{real-world conditions} prompt us to \textcolor{black}{xxx}.
% they do not account for the complete-to-partial ambiguity present in the context of liver surgery.

% \textcolor{blue}{revised here -- mark}
% \textcolor{blue}{

The smoothly shaped surface of the liver produces a challenge for point cloud matching wherein the partial target point cloud may incorrectly match well with multiple candidate regions within the source point cloud due to shallow variations in surface curvature. Although correspondence-based registration methods are well-established in several public natural benchmarks, 
they are not designed to address the complete-to-partial ambiguity inherent in this scenario.

This challenge is similar to a Jigsaw puzzle, where a player tries different candidate spaces to find the correct fit. Inspired by this analogy, we propose converting the complete-to-partial point cloud matching problem into a P2P matching process. By generating several candidate patches from the source point cloud, each roughly the size of the target, and matching them to the target, we can naturally solve the complete-to-partial ambiguity. 
\textcolor{black}{The theoretical formulation can be written as}:

% \begin{equation}
% \textcolor{black}{\{R_k, t_k\} = \arg\min_{R,t} E(S_k, T, R, t),}
% \end{equation}

% \begin{equation}
% \textcolor{red}{(R_k, t_k) = \arg\min_{R,t, S_k \in \mathcal{S}} E(S, T, R, t)}
% \end{equation}

\begin{equation}
\textcolor{black}{(R_k, t_k) = \mathop{\arg\min}_{R,\, t,\, S_k \in \mathcal{S}} E(S_k, T, R, t)}
\end{equation}

\noindent \textcolor{black}{where the complete source point cloud $\mathbf{S} \in \mathbb{R}^{N\times3}$ is represented by several patches  $\mathbf{S}_k \in \mathbb{R}^{M\times3}$ with the same number of points as the target point cloud $\mathbf{T}^{M\times3}$. Here, $E$ is the error function to select the optimal $R,t$ to align the source to the target point cloud.}

\textcolor{black}{Setting the point number of patches serves as a reasonable approximation and simplification, given the prerequisites we further discussed in (see \S\ref{sec:dicsB}).}

\textcolor{black}{The theoretical foundation of our approach is two-fold. First, in comparison to the complete  \(\mathbf{S}\),  a subset \(\mathbf{S}_k\) that closely resembles the shape of \(\mathbf{T}\) is likely to contain distinctive local structures, thereby reducing matching ambiguity. Second, the resulting set of transformations, \(\mathcal{T} = \{(R_1, t_1), (R_2, t_2), ..., (R_k, t_k)\}\), introduces multiple hypotheses, increasing the likelihood of identifying the optimal transformation.} 

\textcolor{black}{The first argument is based on the premise that the ambiguity in complete-to-partial matching is minimized in the limit of the source point cloud approach the same extent and scale as the target point cloud. If the appropriate patch is selected, the registration problem effectively simplifies to a complete-to-complete scenario. The second argument then follows naturally: 
as the number of patches increases and selection is appropriately guided, the probability of identifying the correct transformation correspondingly improves.
% as the number of patches increases and patch selection is guided correctly, the probability of finding the correct transformation also increases.
}

The questions thus become clear:
% The solution involves addressing two key questions: 
\textbf{\ding{182}}~\textit{How do we efficiently generate candidate patches from the source point cloud?} \textbf{\ding{183}}~\textit{How do we select the best transformation?} For consistency in the presentation, we use the point-wise features $\mathbf{x}^{\mathbf{S}}$ and $\mathbf{x}^{\mathbf{T}}$, along with the estimated transformation ${\mathbf{R}_c, \mathbf{t}_c}$ from existing correspondence-based registration methods, as our starting point to address these questions and validate this approach.

\subsection{\textcolor{black}{P2P Module}}
\label{subsec:p2p}

\subsubsection{Visible Source Point Estimation}  \label{subsec:p2p1} In response to \textbf{\ding{182}}, a straightforward approach is to recognize each point from the source point cloud as a patch center and select its nearest neighbors, matching the number of points in the target cloud to generate candidate patches. 
% \textcolor{blue}{
However, this approach requires impractical computational memory. A more efficient approach would be to first narrow down the candidate points before generating these patches, which this part seeks to present rigorously.

% a straightforward approach is to use every point from the source point cloud as a patch center and select their nearest neighbors, matching the number of points in the target cloud, to create candidate patches. 
% However, this method would be computationally insufficient, as registering each point as patch center is impractical.

% \textcolor{blue}{
To narrow down the candidate points, we first design a method to select the top candidates from the source point cloud based on a visibility score, which we define according to a score matrix following \cite{sun2021loftr, li2022lepard} as $S= \mathbf{x}^{\mathbf{S}} (\mathbf{x}^{\mathbf{T}})^\top \in \mathbb{R}^{N\times M}$, where $S$ represents the cosine similarity between source and target features. We then convert this score matrix into visibility scores for the source point cloud by summing over the target dimension: $S_{vis} = \sum_{j=1}^{M} S_{i,j}$. Finally, we select the top $M$ source points with the highest visibility scores to form the visible source point cloud \textcolor{black}{for the next step to generate patch nodes}.

% should now include all the most likely points to match the target point cloud. However, due to complete-to-partial ambiguity, it

% \textcolor{blue}{
  % \textcolor{blue}{

\subsubsection{Patch Candidate Generation}
\label{subsec:p2p2}
The visible source point cloud contains both correct and uncertain points due to complete-to-partial ambiguity. 
% These points are spread across a region, and we assume that a patch within this region can be matched more easily with the target point cloud, offering a more accurate rigid transformation estimation than using the complete point cloud. 
% (see \S\ref{subsubsec:regid_transformation})
The points are distributed across a region, and it is assumed that matching a patch within this region to the target point cloud facilitates a more precise estimation of rigid transformation compared to utilizing the entire point cloud.
Consequently, our next step is to generate a few patches from the region. To do this, we apply farthest point sampling (FPS) \cite{qi2017pointnet} to the visible source point cloud, selecting $k$ points to serve as patch nodes. 

% where we aim to find a patch to align with the target point cloud. Thus, patches are going to be generated from the region  

% Consequently, our next step is to represent the region using a few patches.

\subsubsection{Patch Sampling and Feature Computation} \label{subsec:p2p3} The 3D locations and features of candidate patches are then sampled and generated from the nearest neighbors of these patch nodes, with the same number of points as the target point cloud. We denote the features and 3D locations of the $k$-th source patch as $\mathbf{x}^{\mathbf{S}_k} \in \mathbb{R}^{M \times d}$ and $\mathbf{S}_k  \in \mathbb{R}^{M \times 3}$, respectively.

\subsubsection{Patch-to-Target Matching} \label{subsec:p2p4}
\label{subsubsec:regid_transformation} Each patch candidate generated in the previous step is then matched to the target point cloud to establish correspondences via rigid transformation estimation. Taking one patch as an example, we first compute the score matrix \( S_k = \mathbf{x}^{\mathbf{S}_k} (\mathbf{x}^{\mathbf{T}})^\top \in \mathbb{R}^{\textcolor{black}{M}\times M} \), and apply the dual-softmax operation \cite{sun2021loftr,li2022lepard,rocco2018neighbourhood}:
\begin{equation}
M_k(i,j) = \text{Softmax}(\mathcal{S}_k(i,:)) \cdot \text{Softmax}(\mathcal{S}_k(:,j)),    
\end{equation}
\noindent where the dual-softmax operation converts \( S_k \) into a confidence matrix \( M_k \). Matches are selected from \( M_k \) using the mutual nearest neighbor criterion, where a pair of indices \( (i,j) \) is chosen if its confidence value \( M_k(i,j) \) is the maximum in both \( \mathcal{S}_k(i, \cdot) \) and \( \mathcal{S}_k(\cdot, j) \). 

Once the correspondence set \( \mathcal{C}_k \) is obtained, the weighted SVD is applied to compute the rigid transformations \( \mathbf{R}_k \) and \( \mathbf{t}_k \). The transformations estimated from each patch are inputs to the next step.
% \textcolor{blue}{

% \textcolor{blue}{
% Here, we explain the rationale of our patches-to-partial approach further. If we match the source and target features directly, 
% % without the candidate patch proposal, 
% a point in the partial target point cloud may correspond to multiple similar points in the complete source, 
% which results in the low confidence value of correct matches during the dual-softmax operation. 
% Consequently, the correct matches are more likely to be overlooked under the mutual nearest neighbor criteria. In contrast, our proposed patches-to-partial approach increases the likelihood of detecting correct matches.

% and the rigid transformation that can better align the source point cloud and target point cloud.

% However, a candidate selection procedure is needed to select the most likely rigid transformation from the patches.

% , but it may also introduce additional outliers. To address this, a candidate selection procedure is required, answering \textbf{\ding{183}}.

% These ambiguous points can reduce the 
\subsubsection{Optimal Candidate Selection} \label{subsec:p2p5} A candidate selection procedure is needed in our design to select the most likely rigid transformation from the patches, answering \textbf{\ding{183}}. One possible solution for candidate selection is to apply the inlier number selection rule used in the point-to-node strategy \cite{qin2022geometric, yu2023rotation}. This approach selects the best transformation by choosing the one that results in the most inlier matches:

\begin{equation}
\label{eq:inlier-based}
\mathbf{R} , \mathbf{t} = \max_{\mathbf{R}_i, \mathbf{t}_i} \sum\nolimits_{(p_{j}^S, p_{j}^T) \in \mathcal{C}_{all}} \left[ \lVert \mathbf{R}_j \cdot p_{j}^S \hspace{-3pt} + \mathbf{t}_j - p_{j}^T \rVert_2^2 < \tau \right],
\end{equation}

\noindent where the Iverson bracket \([ \cdot ]\) evaluates whether the condition holds, \(\tau\) is the acceptance radius. \(\mathcal{C}_{all}\) combines all correspondences from the original output and all \(\mathcal{C}_k\) from the previous step. $p_{j}^S$ and $p_{j}^T$ are 3D points within the correspondences.

The transformations \(\mathbf{R}_i\) and \(\mathbf{t}_i\) are from the original output and all candidate transformations \({\mathbf{R}_k, \mathbf{t}_k}\). 
% \textcolor{blue}{
However, this inlier-based selection favors more feature-matching evaluation metrics (e.g., inlier ratio \cite{huang2021predator}), while our goal is to select the best rigid transformation. In light of this view, we propose a straightforward selection rule by directly calculating the mean closest point distance between the transformed source point cloud, \(\mathbf{R}_i\cdot x^S \hspace{-3pt} + \mathbf{t}_i\), and the target point cloud \(x^T\) as:
\begin{equation}
\label{eq:closest}
    \mathbf{R}, \mathbf{t} = \min_{\mathbf{R}_i, \mathbf{t}_i} \sum_{j=1}^{M} \mathbf{Dist}_{min}( \mathbf{R}_i\cdot x^S \hspace{-3pt} + \mathbf{t}_i, x_{j}^T ),
\end{equation}
\noindent where \(\mathbf{Dist}_{min}\) is the closest point operator that computes the distance from \(x_j^T\) to the transformed source point \(\mathbf{R}_i\cdot x^S \hspace{-3pt} + \mathbf{t}_i\).

% \textcolor{blue}{It directly measures how close the transformed source points are $w.r.t.$ to the target point cloud.}

% Parallel quick

% Differentiable Parameter free

% Only parameter

% Different from RANSAC

% As shown in Sec. 4.1, our approach achieves comparable
% registration accuracy with RANSAC but reduces the computation time by more than 100 times. Moreover, unlike
% deep robust estimators [3, 7, 23], our method is parameter-free, and no network training is needed.

\begin{figure*}[h]
\centering
\includegraphics[width=0.95\textwidth]{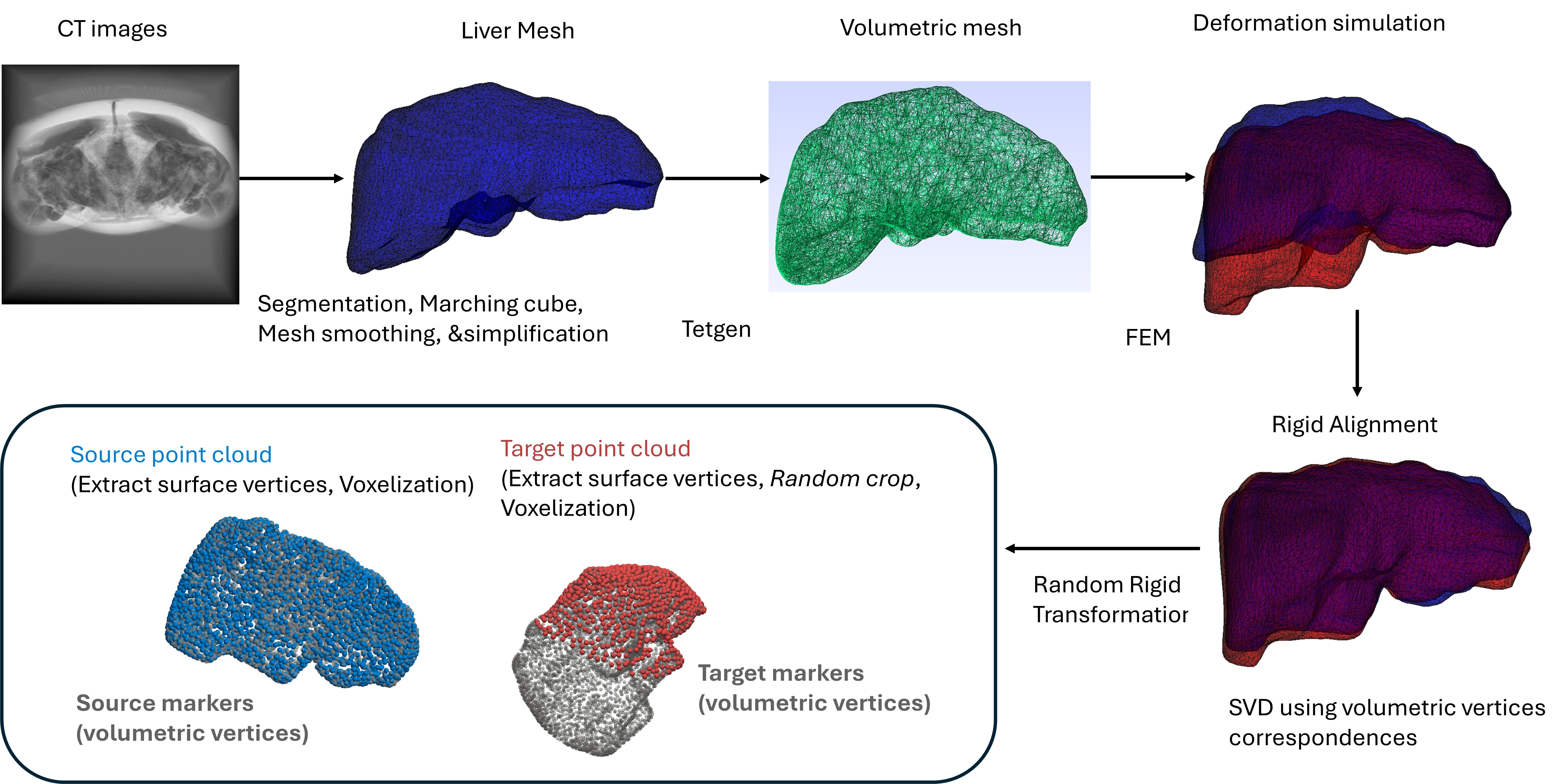}
\caption{Illustration of the \textit{in silico} phantom generation process. Source and target point clouds/ meshes are shown in blue and red, respectively.}
\label{fig:simulation}
\end{figure*}

\begin{figure*}[h]
\centering
\includegraphics[width=0.99\textwidth]{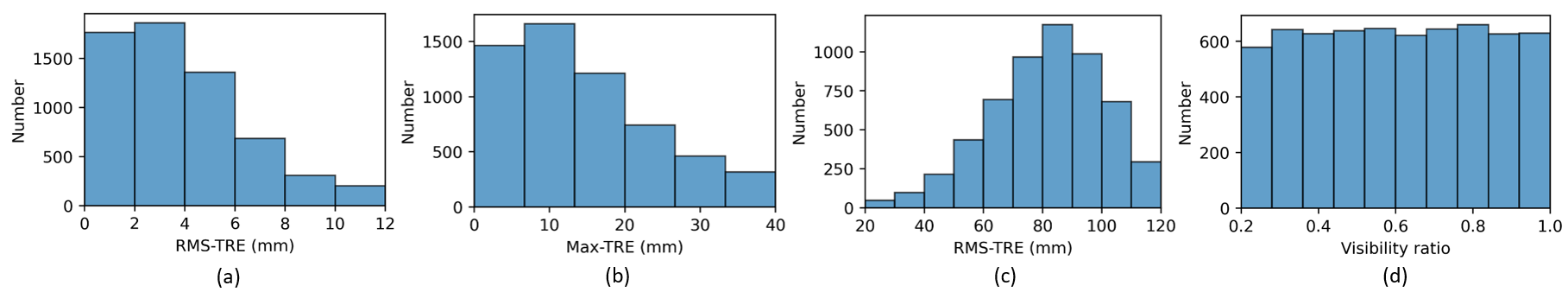}
\caption{\textcolor{black}{Properties of the testing set from the \textit{in silico} phantom dataset: (a) Distribution of RMS-TRE across all sample pairs after rigid alignment using volumetric vertices correspondences to remove rigid components. (b) Distribution of max-TRE per sample pair after rigid alignment. (c) Initial RMS-TRE across samples after applying random rigid transformations, used to evaluate rigid registration methods. (d) Visibility ratio of the target point cloud across samples after random cropping.}}
\label{fig:vis_def}
\end{figure*}

\section{Experiments}

\subsection{Datasets and Evaluation}
\label{sec:dataset}

We use the \textit{in silico} dataset for both training and testing, while the \textit{in vitro} dataset is used exclusively for testing. The same procedure (see \S\ref{subsec:Preprocessing}) is applied to preprocess the input data for the baseline and the proposed methods (see \S\ref{subsec:baselines}).

\subsubsection{\textit{In silico} phantom dataset} \label{subsubsec:in_silico_dataset} Fig. \ref{fig:simulation} illustrates the generation process of the \textit{in silico} dataset. We utilized CT scan volumes from the training dataset of the Medical Segmentation Decathlon \cite{antonelli2022medical} to generate liver models represented by $131$ triangular meshes. These meshes were extracted using the marching cubes algorithm from the scikit-image Python library \cite{van2014scikit}. To enhance the quality of the mesh, we applied a Laplacian smoothing kernel and used vertex clustering with a voxel size of $2$ mm from the Open3D Python library \cite{Zhou2018}. Finally, we used Tetgen \cite{hang2015tetgen} to generate volumetric mesh from the surface meshes. The liver models were split into $120$ models for training and 11 models for testing. Each liver model was randomly scaled in the x, y, and z dimensions to augment the data. The isotropic scale factors were uniformly generated from the range [$0.5$, $1$]. This process generated $10$ additional liver models for each original liver, resulting in $1,320$ liver models for training and $121$ liver models for testing. 

For each liver model, we generated $10$ deformations using the pipeline proposed in \cite{pfeiffer2020non} via a finite element model \textcolor{black}{(neo-Hookean model)} and used its default parameters. Some simulated deformations feature rigid transformations instead of deformation when the randomly generated zero boundary condition is small. To eliminate rigid transformation, we further align the undeformed and deformed models with the SVD using the volumetric vertices as fiducial markers. After the alignment, the average root mean square error of the fiducial markers is $3.37$ mm, with a maximum value of $11.38$ mm, which is similar to the deformation range in Sparse Non-rigid Registration Challenge dataset~\cite{heiselman2024image}.

For both training and testing datasets, we followed the previous 
works \cite{huang2021predator,li2022lepard,yang2023learning} to perform random cropping and rigid transformations on the deformed surface for each deformation. We calculate the visibility ratio as  $M/N$. The visibility of the target point clouds was between $0.2$ and $1.0$. Regarding the rigid transformations, we sample rotations according to three Euler angle rotations in the range $[0, 2\pi]$ and translations in the range [$-100$, $100$] mm.

For the testing dataset, we generated five target point clouds for each deformation. Consequently, the testing dataset contained $121$ source point clouds with $6,050$ target point clouds.  \textcolor{black}{The deformation and visibility of the testing dataset are shown in Fig. \ref{fig:vis_def}.}

The training dataset comprised $1,320$ liver models with $13,200$ deformed liver models after the deformation simulation. A specialized training strategy will be introduced later in \ref{sec:Implementations} to utilize the training dataset.

As illustrated in Fig. \ref{fig:simulation}, the volumetric vertices are assumed to be fiducial markers for measuring registration accuracy.

% The source point clouds were extracted from undeformed liver surfaces, while the

% resulting in each source point cloud having 50 target point clouds. 

% Consequently, the training dataset comprised 1320 liver models with 13200 deformed liver models

% , while the testing dataset contained 121 source point clouds with 6050 target point clouds. 

% \begin{figure}[!t]
%     \centering
%     \includegraphics[width=1\linewidth]{phantom.JPG}
%     \caption{\textcolor{black}{redo}.
%     }
%     \label{fig:phantom}
% \end{figure}

\subsubsection{\textit{In vitro} phantoms}

We used the phantom dataset constructed in \cite{yang2024boundary} as a baseline to create another testing dataset to prove the generality of our proposed method. The phantom dataset contains four undeformed (source) and deformed (target) models with known embedded fiducial markers as validation points to compute the target registration errors. CT scans of both undeformed and deformed phantoms were acquired, and the ground truth surface mesh and fiducial landmarks were manually segmented from the CT scans. We generated $200$ target point clouds from the complete target model surface, producing $800$ testing samples.

% The centroids of the fiducial landmarks were used as their locations. Correspondences of fiducial landmarks were also manually identified. After eliminating the rigid part, the mean Target Registration Errors (TREs) ranged from around 3 to 5 mm, and the maximum TRE values ranged from around 10 to 24 mm. One example is illustrated in Fig. \ref{fig:phantom}. Same as the \textit{in silico} data, we generate 200 target point clouds from the complete target model surface, resulting in 800 testing samples.

% \subsubsection{\textit{In vivo} Liver with Breathing Deformation}

\subsubsection{Evaluation} A large number of evaluation metrics \cite{huang2021predator, li2022lepard, xu2021omnet} exist to measure point cloud registration. However, these metrics rely on ground truth rigid transformations or one-to-one correspondence, which do not apply to our settings. Also, they do not provide insight into subsurface registration errors. 

Here, we use the root mean square target registration error (RMS-TRE) to measure the final outcome of the registration methods on the above datasets. RMS-TRE offers an unbiased assessment of registration accuracy across the liver, including fiducial markers underneath the surface, and directly reflects the precision required when overlaying preoperative data onto intraoperative data. It takes the form as:
\begin{equation}
\label{eq:rms-tre}
\text{RMS-TRE}=\sqrt{\frac{\lVert \mathbf{Y} - \mathbf{W}(\mathbf{X})  \rVert_{2}^2}{N_f}},
\end{equation}
\noindent where $\mathbf{Y}$ and $\mathbf{X}$ represent the locations of fiducial markers in the source and target, respectively, and $N_f$ is the number of fiducial markers.

\subsection{Implementations}
\label{sec:Implementations}

\subsubsection{Preprocessing} \label{subsec:Preprocessing} The preprocessing step is crucial as the point cloud feature learning models typically require a consistent point cloud density to extract features effectively. We begin by centering the source point cloud by subtracting its centroid, then scale it to fit within a unit sphere using the maximum distance from the origin. The target point cloud is processed similarly but scaled using the same factor as the source. Both point clouds are then voxelized with a voxel size of $0.04$. 

% An alternative preprocessing method, FPS, reduces the point cloud to a fixed number of points. However, FPS is sensitive to outliers and irregular points, making voxelization a potentially more reliable approach. 

\subsubsection{Baselines} \label{subsec:baselines} We used the official implementations of the following methods: (1). Go-ICP \cite{yang2015go}, a traditional method that globally searches for a minimizing condition across SE(3). (2). RegTr \cite{yew2022regtr}, an end-to-end point cloud correspondence network that predicts a final set of correspondences directly rather than matching source and target features. (3). RoITr \cite{yu2023rotation}, a state-of-the-art method that utilizes a point-to-node strategy. (4). Lepard \cite{li2022lepard}, another state-of-the-art method, excelling in rigid and non-rigid point cloud registration datasets. (5). LiverMatch \cite{yang2023learning}, an available correspondence-based point cloud registration method specifically developed for liver surgery. RegTr \cite{yew2022regtr}, Lepard \cite{li2022lepard}, and LiverMatch \cite{yang2023learning} leverage the power of KPConv\cite{thomas2019kpconv} in their point cloud feature learning. The initial cell size in KPConv, which controls the downsampling rate, is set to $0.02$ to enable gradual downsampling on our datasets. All the learning-based methods are implemented in PyTorch~\cite{NEURIPS2019_9015}.

\subsubsection{Proposed module} The only hyperparameter of the proposed module is the
number of candidate patches, K, whose default value is set to $5$. In \S\ref{subsec:ablation},
a sensitivity study is conducted to evaluate the impact of this hyperparameter on the downstream performance of the registration methods.

\subsubsection{Procrustes} In addition to the baseline methods, we also include Procrustes, which uses SVD with ground truth fiducial markers to estimate the rigid registration. This method serves as a reference for deformation and potentially offers the best rigid registration that baseline methods can achieve.

\subsubsection{Statistical hypothesis test}\label{stat} The Wilcoxon Rank Sum Test at a significance level of $\alpha = 0.05$ is used to assess whether integrating the proposed module with a baseline method significantly improves registration results compared to the original baseline performance.

\subsubsection{Training} We used a special training strategy that fully exploits the \textit{in silico} training dataset by dynamically selecting source and target point cloud pairs during training. Specifically, two source and target point clouds are randomly chosen from a liver and its deformations within the training dataset. Random cropping and rigid transformations are then applied to the target point cloud, as described in \S\ref{subsubsec:in_silico_dataset}. Some baselines required either rigid transformation or correspondences between source and target point cloud. The required rigid transformation is obtained from Procrustes, and the correspondences are determined using nearest neighbor searching between the source and target point clouds within a radius of $0.04$ after applying the Procrustes. For all methods, the maximum number of training epochs is set to $150$, and the batch size is set to $1$. Training is performed on an NVIDIA A100 GPU, with Lepard taking approximately three days to train, while other methods require about one day.

\section{Results}

\subsection{\textit{In Silico} Phantom Validation}
\label{sec:in_silico}

\begin{table*}[htb!]%[!htpb]
	\centering
	\caption{ Comparison of registration errors on the \textit{In Silico} phantom dataset across different visibility ratios. The average RMS-TRE ± standard deviation is reported in millimeters.
 }
	\label{tab:vis_sim}
	\resizebox{1.0  \linewidth}{!}{
    \begin{tabular}{lcccccccc}
			\hline
   \hline
            %\hline
            Visibility ratio   & $[0.2, 0.3)$ 
            & $[0.3, 0.4)$ 
            & $[0.4, 0.5)$ 
            & $[0.5, 0.6)$ 
            & $[0.6, 0.7)$ 
            & $[0.7, 0.8)$ 
            & $[0.8, 0.9)$ 
            & $[0.9, 1]$\\

                \hline
Procrustes
           
&        3.19 $\pm$ 1.93 
 
& 3.44 $\pm$ 2.16 
 
& 3.41 $\pm$ 2.13 
 
& 3.42 $\pm$ 2.19 
 
& 3.38 $\pm$ 2.17 
 
& 3.41 $\pm$ 2.17 
 
& 3.39 $\pm$ 2.01 
 
& 3.31 $\pm$ 2.08 

                \\ 		
\hline
  GO-ICP \cite{yang2015go}
 
       & 66.96 $\pm$ 17.2 

& 66.98 $\pm$ 19.52 

& 56.4 $\pm$ 23.52 

& 55.61 $\pm$ 25.41 

& 44.83 $\pm$ 29.27 

& 33.56 $\pm$ 29.94 

& 19.9 $\pm$ 24.21 

& 5.42 $\pm$ 2.65 
    
   \\
                  RoITr \cite{yu2023rotation}
                  
& 20.21 $\pm$ 25.12 
 
& 10.77 $\pm$ 12.71 
 
& 7.24 $\pm$ 8.13 
 
& 5.6 $\pm$ 6.05 
 
& 4.61 $\pm$ 3.57 
 
& 4.19 $\pm$ 2.61 
 
& 3.91 $\pm$ 2.36 
 
& 3.75 $\pm$ 2.46 
   \\

RegTr \cite{yew2022regtr}
              
& 84.27 $\pm$ 29.46 
 
& 69.89 $\pm$ 30.16 
 
& 59.66 $\pm$ 30.45 
 
& 48.08 $\pm$ 30.11 
 
& 33.53 $\pm$ 24.36 
 
& 20.24 $\pm$ 11.89 
 
& 12.05 $\pm$ 4.16 
 
& 5.82 $\pm$ 2.41           
\\
\hline
LiverMatch \cite{yang2023learning}
& 12.85 $\pm$ 13.87 
 
& 8.09 $\pm$ 6.88 
 
& 6.65 $\pm$ 5.15 
 
& 5.65 $\pm$ 3.98 
 
& 4.79 $\pm$ 2.82 
 
& 4.24 $\pm$ 2.45 
 
& 3.84 $\pm$ 2.09 
 
& 3.56 $\pm$ 2.07 
 \\
LiverMatch\cite{yang2023learning} + proposed P2P

& \textbf{8.74 $\pm$ 10.23} 
 
& \textbf{6.55 $\pm$ 5.54} 
 
& \textbf{5.5 $\pm$ 3.69} 
 
& \textbf{5.00 $\pm$ 3.30} 
 
& \textbf{4.40 $\pm$ 2.71} 
 
& \textbf{4.08 $\pm$ 2.46} 
 
& \textbf{3.78 $\pm$ 2.15} 
 
& 3.58 $\pm$ 2.16       
               \\
               \hline
                Lepard \cite{li2022lepard}
                 
& 9.47 $\pm$ 14.74 
 
& 6.17 $\pm$ 4.83 
 
& 5.22 $\pm$ 3.7 
 
& 4.76 $\pm$ 4.33 
 
& 4.20 $\pm$ 2.58 
 
& 3.96 $\pm$ 2.43 
 
& 3.74 $\pm$ 2.17 
 
& 3.57 $\pm$ 2.22

               \\     
Lepard \cite{li2022lepard} + proposed P2P
               
& \textbf{6.73 $\pm$ 5.96} 
 
&  \textbf{5.63 $\pm$ 3.72} 
 
&  \textbf{4.96 $\pm$ 3.28} 
 
&  \textbf{4.60 $\pm$ 2.93} 
 
&  4.22 $\pm$ 2.68
 
&  4.00 $\pm$ 2.50 
 
&  3.78 $\pm$ 2.22 
 
&  3.59 $\pm$ 2.25 
               
\\              
                             
 			\hline
    \hline
	\end{tabular}
	}
\end{table*}

\begin{table}[htb!]%[!htpb]
	\centering
	\caption{Registration errors on the \textit{in silico} phantom dataset at different noise levels, within the visibility ratio range [0.2, 0.3). The average RMS-TRE ± standard deviation is reported in millimeters.}
	\label{tab:noise}
	\resizebox{0.95\linewidth}{!}{
    \begin{tabular}{lccc}
			\hline
   \hline
            %\hline
            Noise level & 0 mm
            & 2 mm
            & 4 mm
           
            \\
            \hline
            Procrustes
           &  3.37 $\pm$ 2.11 
           &  3.37 $\pm$ 2.11 
           &  3.37 $\pm$ 2.11 
            \\
           
%   GO-ICP \cite{yang2015go}
%     & 
%    & 51.0 $\pm$ 39.42 
%    & 51.5 $\pm$ 39.38 
%    & 52.1 $\pm$ 39.24 
% \\
% \hline
% RegTr \cite{yew2022regtr}
% &
% & 41.32 $\pm$ 35.22 

%                  \\ 	          
 			\hline
              
 			LiverMatch \cite{yang2023learning}
         & 12.85 $\pm$ 13.87 
         & 13.72 $\pm$ 15.48 
              & 21.85 $\pm$ 23.2            
\\
LiverMatch \cite{yang2023learning}  + proposed P2P
& \textbf{8.74 $\pm$ 10.23} 
& \textbf{9.19 $\pm$ 9.48} 
& \textbf{14.58 $\pm$ 18.11} 
\\

%                 \hline             		
%                 Lepard \cite{li2022lepard}
%                 & \multirow{2}{*}{ RANSAC ICP}
% & 8.07 $\pm$ 12.16 
% & 8.52 $\pm$ 12.38 
% & 12.42 $\pm$ 21.51 
%                 \\
%  			LiverMatch \cite{yang2023learning}
%     & 
%  & 9.44 $\pm$ 9.68 
% & 11.07 $\pm$ 13.78 
% & 17.93 $\pm$ 24.2 
%                   \\

\hline  
  Lepard \cite{li2022lepard}
                
             & 9.47 $\pm$ 14.74 
             & 10.07 $\pm$ 15.28 
             & 14.7 $\pm$ 21.02 
               \\

Lepard \cite{li2022lepard} + proposed P2P
& \textbf{6.73 $\pm$ 5.96} 
& \textbf{7.22 $\pm$ 8.45} 
& \textbf{9.85 $\pm$ 14.88}

\\
 			\hline
    \hline
	\end{tabular}
	}
\end{table}

\begin{table}[htb!]%[!htpb]
	\centering
	\caption{\textcolor{black}{Registration errors on the \textit{in silico} phantom dataset at different deformation ranges, within the visibility ratio range [0.2, 0.3). The average RMS-TRE ± standard deviation is reported in millimeters.}}
	\label{tab:deform_tre}
	\resizebox{0.95\linewidth}{!}{
    \begin{tabular}{lcc}
			\hline
   \hline
            %\hline
            Deformation range (RMS-TRE)   & $(0, 6)$ 
            & $[6, 12]$ \\

        \hline
            Procrustes
            & 2.82 $\pm$ 1.51 
 
& 7.30 $\pm$ 1.25 

                 \\ 	          
 			\hline
              
 			LiverMatch \cite{yang2023learning}
& 12.54 $\pm$ 13.75 
 
& 17.24 $\pm$ 15.93 
\\
LiverMatch \cite{yang2023learning} + proposed P2P
& \textbf{8.54 $\pm$ 11.16} 
 
& \textbf{12.54 $\pm$ 5.13} 
              
               \\
\hline  
  Lepard \cite{li2022lepard}
& 8.92 $\pm$ 14.66 
 
& 15.50 $\pm$ 13.82 
               
               \\

Lepard \cite{li2022lepard} + proposed P2P
& \textbf{6.19 $\pm$ 5.50} 
 
& \textbf{12.85 $\pm$ 7.24}

\\
 			\hline
    \hline
	\end{tabular}
	}
\end{table}

\begin{figure*}[!t]
    \centering
\includegraphics[width=0.95\linewidth]{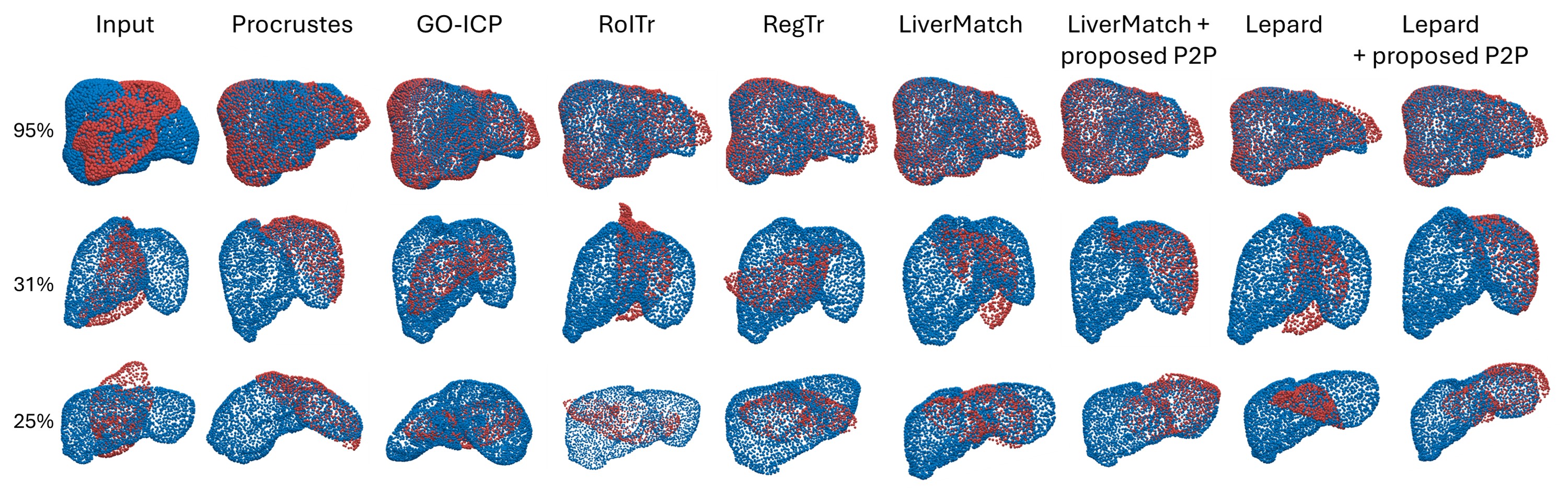}
    \caption{Qualitative comparison of registration results on the \textit{in silico} phantom dataset. The source and target point clouds are shown in blue and red, respectively.
    } \label{fig:qualitative_silico}
    % \vspace{-1.3em}
\end{figure*}

\begin{figure}[htb!]
    \centering
    \includegraphics[width=0.85\linewidth]{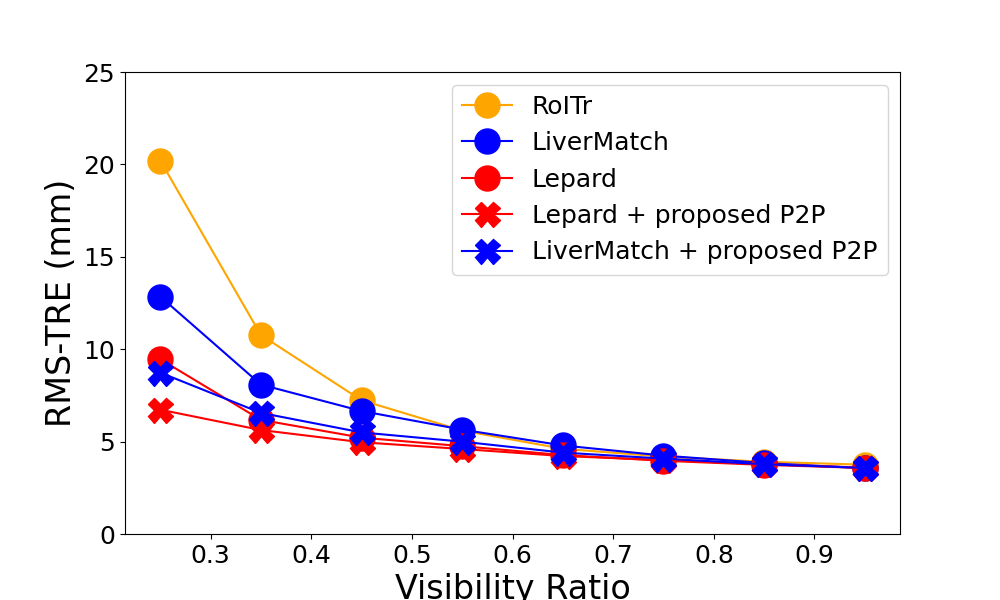}
    \caption{Comparison of registration errors from the \textit{in silico} phantom dataset, with the average RMS-TRE values plotted against different visibility ratios using the bin centers from Table \ref{tab:vis_sim}.
    }
     % \vspace{-1.5em}
    \label{fig:Compare_sim_curves}
\end{figure}

\subsubsection{Performance against Baselines} We begin by evaluating the performance of the baselines under varying visibility ratios, with the quantitative results summarized in Table \ref{tab:vis_sim}.

At higher visibility ratios (i.e., visibility $\in [0.9, 1.0]$), the baseline methods perform similarly to Procrustes, where rigid registration achieves the lowest errors, even in the presence of deformation. However, as the visibility ratio decreases, the performance of GO-ICP declines sharply, followed by RegTr, RoITr, LiverMatch, and Lepard. Among these, Lepard shows better robustness to changes in visibility, although its RMS-TRE still increases noticeably from 3.56 mm at visibility ratios $\in [0.9, 1.0]$ to 9.47 mm at visibility ratios $\in [0.2, 0.3]$

We then examine the results obtained by Lepard and LiverMatch with our proposed module, alongside those from RoITr quantitatively and qualitatively in Table~\ref{tab:vis_sim} and 
% , as shown in 
Fig. \ref{fig:Compare_sim_curves}, respectively. 
At the lowest visibility ratio range (i.e., visibility $\in [0.2, 0.3]$), the registration errors for Lepard decrease from $9.47$ mm to $6.73$ mm, and for LiverMatch, from $12.85$ mm to $8.74$ mm, following the integration of our proposed module. \textcolor{black}{These improvements over baseline results are confirmed to be statistically significant ($p<0.05$), based on the statistical analysis method described in \S\ref{stat}.}

\subsubsection{Sensitivity to Noise} 

Given the promising results achieved by Lepard and LiverMatch with our proposed method, we further investigate their performance in the presence of noise and across groups with varying deformations in Table \ref{tab:noise}.
% we evaluate the robustness of the methods under different noise conditions. 
In our setting, noise is generated by sampling uniform random values from the range $[-0.5, 0.5]$, scaling them by the noise level, and applying them to the target point cloud. When the noise level increases from $2$ to $4$ mm, the performances of LiverMatch and Lepard decline by $59.25\%$ and $43.92\%$, respectively. 
Our proposed method, on the other hand, helps both methods maintain lower RMS-TRE at these noise levels.

% For example, at a noise level of $4$ mm, the errors are reduced from $12.42$ mm to $9.95$ mm for Lepard and from $21.85$ mm to $14.58$ mm for LiverMatch.

\subsubsection{\textcolor{black}{Sensitivity to Deformation}} 

\textcolor{black}{Since rigid registration methods are expected to provide a reasonable initialization for subsequent non-rigid registration,} \textcolor{black}{we also examine their robustness under varying deformation levels and evaluate the effectiveness of the proposed P2P module.} \textcolor{black}{As shown in Table~\ref{tab:deform_tre}, the P2P module consistently enhances registration performance across different deformation ranges within the low visibility ratio \([0.2, 0.3)\).} \textcolor{black}{Specifically, for mild deformation \([0, 6)\), registration errors are reduced from 12.54 mm to 8.54 mm for LiverMatch and from 8.92 mm to 6.19 mm for Lepard.} \textcolor{black}{Under more severe deformation \([6, 12]\), errors decrease from 17.24 mm to 12.54 mm for LiverMatch and from 15.50 mm to 12.85 mm for Lepard.} \textcolor{black}{These results demonstrate the robustness of the proposed P2P strategy in handling both low visibility and large deformation scenarios.}

\subsubsection{\textcolor{black}{Success Rate}}

\textcolor{black}{To determine the success rate of this automatic initial rigid registration approach, we define the success rate as the percentage of examples for which the RMS-TRE is below a certain threshold $\tau$.}

\begin{figure}[thb!]
    \centering
\includegraphics[width=0.75\linewidth]{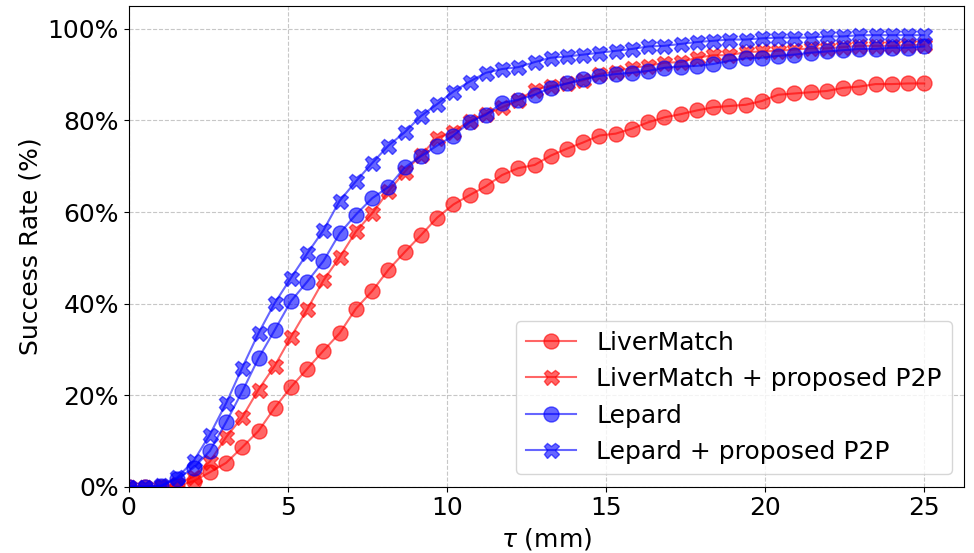}
    \caption{\textcolor{black}{Comparison of success rates at different thresholds ($\tau$) on the \textit{in silico} phantom dataset within the visibility ratio range [0.2, 0.3).}
    }
    \label{fig:success_rate}
\end{figure}

\textcolor{black}{In Fig. \ref{fig:success_rate}, we further demonstrate the effectiveness of the P2P module by comparing the success rate curves of LiverMatch and Lepard, both with and without the integration of P2P. Success rates achieved with P2P are always higher than results without P2P.}

\textcolor{black}{
We also evaluate robustness using the success rate (\(\tau = 20\) mm) under different deformation levels. As shown in Table~\ref{tab:deform_SR}, the P2P module increases the success rate by 11.2\% for LiverMatch and by 4.2\% for Lepard under mild deformation (0-6 mm) and by 8.3\% and 5.0\%, respectively, under more severe deformation (6-12 mm).} 

\textcolor{black}{These results further confirm the robustness and effectiveness of P2P in enhancing registration performance under challenging conditions.}

\begin{table}[htb!]%[!htpb]
	\centering
	\caption{\textcolor{black}{Registration success rate ($\tau =20 \ mm$) in percentage on the \textit{in silico} phantom dataset at different deformation ranges \textcolor{black}{ measured by RMS-TRE in millimeters}, within the visibility ratio range [0.2, 0.3).}}
	\label{tab:deform_SR}
	\resizebox{0.75\linewidth}{!}{
    \begin{tabular}{lcc}
			\hline
   \hline
            %\hline
            Deformation range (RMS-TRE)    & $(0, 6)$ 
            & $[6, 12]$ 
            
            \\
            \hline
            Procrustes
            & 100.00 
            & 100.00
                 \\ 	          
 			\hline
              
 			LiverMatch \cite{yang2023learning}
& 84.38
 
& 83.33 
\\
LiverMatch \cite{yang2023learning} + proposed P2P
& \textbf{95.54}
 
& \textbf{91.67} 
              
               \\
\hline  
  Lepard \cite{li2022lepard}
&  94.49
 
& 86.67
               
               \\

Lepard \cite{li2022lepard} + proposed P2P
&  \textbf{98.66}
 
&  \textbf{91.67}

\\
 			\hline
    \hline
	\end{tabular}
	}
\end{table}

\subsubsection{Visual Assessment}

We further visualize the results in Fig. \ref{fig:qualitative_silico}. In the first row, we show cases with a high visibility ratio, where all methods produce reasonable results. However, in the second and last rows, when the visibility ratio is low, all methods struggle to align 
the source to the target point cloud properly. 
After incorporating our proposed module, both LiverMatch and Lepard achieved reasonable results.

\begin{figure}[thb!]
    \centering
    \includegraphics[width=0.75\linewidth]{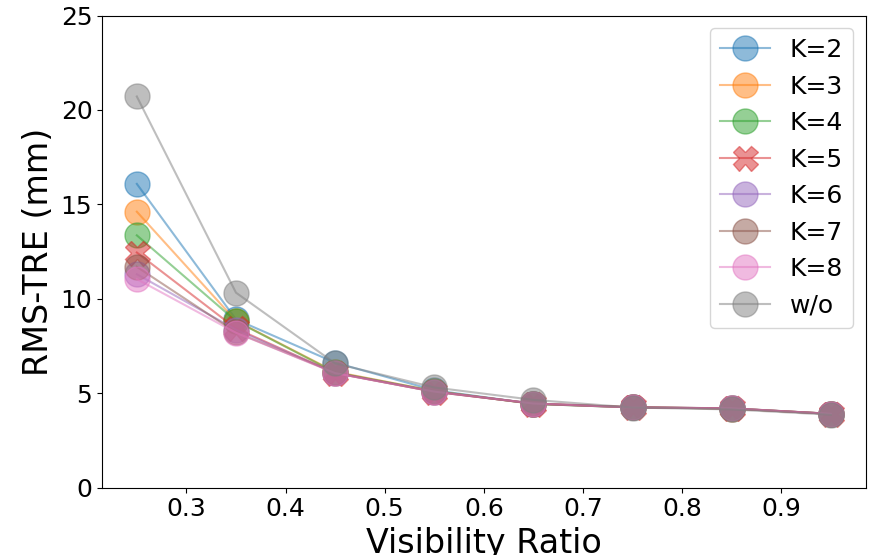}
    \caption{Sensitivity study on patch number K, performed on the \textit{in vitro} phantom dataset at different visibility ratios, using Lepard as the baseline for the proposed module. 
    }
    \label{fig:ab_K}
\end{figure}

\begin{figure}[thb!]
    \centering
    \includegraphics[width=0.85\linewidth]{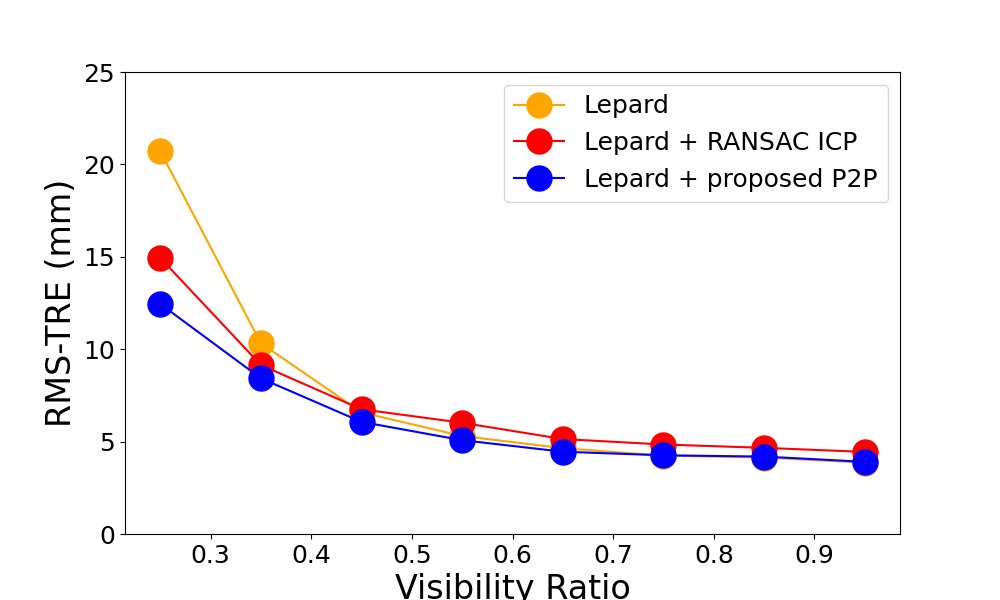}
    \caption{Comparison of the proposed module with RANSAC ICP on the \textit{in vitro} phantom dataset.
    }
    \label{fig:RANSAC_ours}
\end{figure}

\begin{table}[thb!]%[!thpb]
	\centering
	\caption{Ablation study on the selection method and candidate proposal, performed on the \textit{in vitro} phantom dataset at a visibility range of [0.2, 0.3), using Lepard as the baseline for the proposed module. The average RMS-TRE ± standard deviation is reported in millimeters.}
	\label{tab:ablation}
\resizebox{0.95\linewidth}{!}{	
\begin{tabular}{ll|ccc}
 \hline
 \hline
% \multicolumn{2}{l|}{Ablation Target}  
% & RE\\         
% \hline
\multirow{2}{*}{Selection rule}  
& Inlier number 
& 14.14 $\pm$ 16.70 
\\
& Closest point distance (default)  
& \textbf{12.45 $\pm$ 15.72}
\\
\hline
\multirow{2}{*}{Candidate proposal}  
& w/o 
& 14.91 $\pm$ 19.06 
\\
& w/  (default)
&  \textbf{12.45 $\pm$ 15.72} 
\\
\hline
% \multirow{2}{*}{Additional network}  
% & w/o (default)
% & 
% \\
% & w/ vanilla Transformer
% & 
% &   
% &  
% \\
% \hline
% \hline
\end{tabular}
}
% \vspace{-1.5em}
\end{table}

\subsubsection{Running Time Analysis}: \textcolor{black}{Table \ref{tab:runtime} summarizes the average running time of each registration method on the \textit{in silico} phantom dataset. Traditional optimization-based GO-ICP is significantly slower than learning-based approaches, with an average time of 89.62 seconds per case. In contrast, learning-based methods complete registration within fractions of a second. The P2P module introduces only a slight increase in runtime while providing improved robustness to partial visibility, as previously demonstrated in Table \ref{tab:vis_sim}.}

\begin{table}[htbp]
\centering
\caption{\textcolor{black}{Running time comparison of different registration methods on the \textit{In Silico} phantom dataset.}}
\begin{tabular}{lc}
\hline
Method & Averaged Running Time (s) \\
\hline
GO-ICP                & 89.62 \\
\hline
RoITr                 & 0.25 \\
RegTr                 & 0.12\\
\hline
LiverMatch            & 0.07\\
LiverMatch + P2P      & 0.10\\
\hline
Lepard                & 0.10\\
Lepard + P2P          & 0.12\\
\hline
\end{tabular}
\label{tab:runtime}
\end{table}

\subsection{\textit{In Vitro} Phantom Validation}
\label{sec:phantoms}

\begin{table*}[htb!]%[!htpb]
	\centering
	\caption{Comparison of registration errors on the \textit{in vitro} phantom dataset across different visibility ratios. The average RMS-TRE ± standard deviation is reported in millimeters.}
	\label{tab:vis_phantom}
	\resizebox{1.0\linewidth}{!}{
    \begin{tabular}{lccccccccc}
			\hline
   \hline
            %\hline
            Visibility ratio & $[0.2, 0.3)$ 
            & $[0.3, 0.4)$ 
            & $[0.4, 0.5)$ 
            & $[0.5, 0.6)$ 
            & $[0.6, 0.7)$ 
            & $[0.7, 0.8)$ 
            & $[0.8, 0.9)$ 
            & $[0.9, 1)]$\\
            \hline
            Procrustes
 & 3.66 $\pm$ 1.69 
 
& 3.67 $\pm$ 1.78 
 
& 3.91 $\pm$ 1.74 
 
& 3.58 $\pm$ 1.8 
 
& 3.43 $\pm$ 1.68 
 
& 3.62 $\pm$ 1.7 
 
& 3.77 $\pm$ 1.6 
 
& 3.65 $\pm$ 1.91

            \\
\hline
  GO-ICP \cite{yang2015go}
 & 87.15 $\pm$ 25.91 
             
                & 82.58 $\pm$ 25.45 
                 
                & 75.34 $\pm$ 25.53 
                 
                & 72.2 $\pm$ 30.03 
                 
                & 61.89 $\pm$ 30.91 
                 
                & 47.04 $\pm$ 36.66 
                 
                & 21.16 $\pm$ 26.16 
                 
                & 9.31 $\pm$ 19.08 
   
\\

RoITr \cite{yu2023rotation}
 & 33.8 $\pm$ 24.06 

& 17.03 $\pm$ 17.2 

& 12.67 $\pm$ 14.43 

& 8.34 $\pm$ 9.33 

& 5.7 $\pm$ 3.69 

& 4.66 $\pm$ 2.21 

& 4.39 $\pm$ 1.8 

& 4.1 $\pm$ 2.09 

\\

RegTr \cite{yew2022regtr}
& 66.25 $\pm$ 16.15 

& 55.91 $\pm$ 19.19 

& 45.02 $\pm$ 20.19 

& 37.39 $\pm$ 21.67 

& 29.74 $\pm$ 20.48 

& 16.15 $\pm$ 13.07 

& 8.76 $\pm$ 2.66 

& 5.16 $\pm$ 1.8 

 \\
 \hline
LiverMatch \cite{yang2023learning}
& 20.54 $\pm$ 22.33 

& 12.46 $\pm$ 12.87 

& 8.26 $\pm$ 5.34 

& 6.69 $\pm$ 4.29 

& 5.26 $\pm$ 3.16 

& 4.66 $\pm$ 2.58 

& 4.18 $\pm$ 1.7 

& 3.85 $\pm$ 1.85   
\\

 			LiverMatch \cite{yang2023learning} + proposed P2P

      & \textbf{14.97 $\pm$ 18.73} 

& \textbf{9.21 $\pm$ 7.95} 

& \textbf{6.98 $\pm$ 4.51} 

& \textbf{5.66 $\pm$ 2.80} 

& \textbf{4.74 $\pm$ 3.03} 

& \textbf{4.5 $\pm$ 2.26} 

& \textbf{4.17 $\pm$ 1.74} 

& \textbf{3.83 $\pm$ 1.92}    \\
                         
%                \\
%                 \hline             		
%                 Lepard \cite{li2022lepard}
%                 & \multirow{2}{*}{RANSAC ICP}
% & 14.94 $\pm$ 18.37 

% & 9.14 $\pm$ 7.88 

% & 6.75 $\pm$ 2.97 

% & 6.02 $\pm$ 2.65 

% & 5.14 $\pm$ 2.11 

% & 4.84 $\pm$ 1.9 

% & 4.66 $\pm$ 1.69 

% & 4.44 $\pm$ 1.92 

%                 \\
%  			LiverMatch \cite{yang2023learning}
%     & 
%  & 15.34 $\pm$ 18.52 

% & 9.71 $\pm$ 9.44 

% & 7.5 $\pm$ 3.86 

% & 6.45 $\pm$ 2.84 

% & 5.74 $\pm$ 2.52 

% & 5.39 $\pm$ 1.97 

% & 5.27 $\pm$ 1.77 

% & 4.87 $\pm$ 1.88 
%                    \\

\hline       
Lepard \cite{li2022lepard}
& 20.71 $\pm$ 23.39 

& 10.32 $\pm$ 12.03 

& 6.58 $\pm$ 4.33 

& 5.3 $\pm$ 2.62 

& 4.65 $\pm$ 3.09 

& 4.24 $\pm$ 1.96 

& 4.14 $\pm$ 1.72 

& 3.86 $\pm$ 1.96 
\\
          		
Lepard \cite{li2022lepard} + proposed P2P

& \textbf{12.45 $\pm$ 15.72} 

& \textbf{8.43 $\pm$ 9.64} 

& \textbf{6.07 $\pm$ 3.12} 

& \textbf{5.08 $\pm$ 2.58} 

& \textbf{4.45 $\pm$ 2.38} 

& 4.26 $\pm$ 2.01 

& 4.19 $\pm$ 1.78 

& 3.91 $\pm$ 2.02 
\\
\hline
\hline
	\end{tabular}
	}
\end{table*}

\begin{figure*}[htb!]
    \centering
    \includegraphics[width=0.95\linewidth]{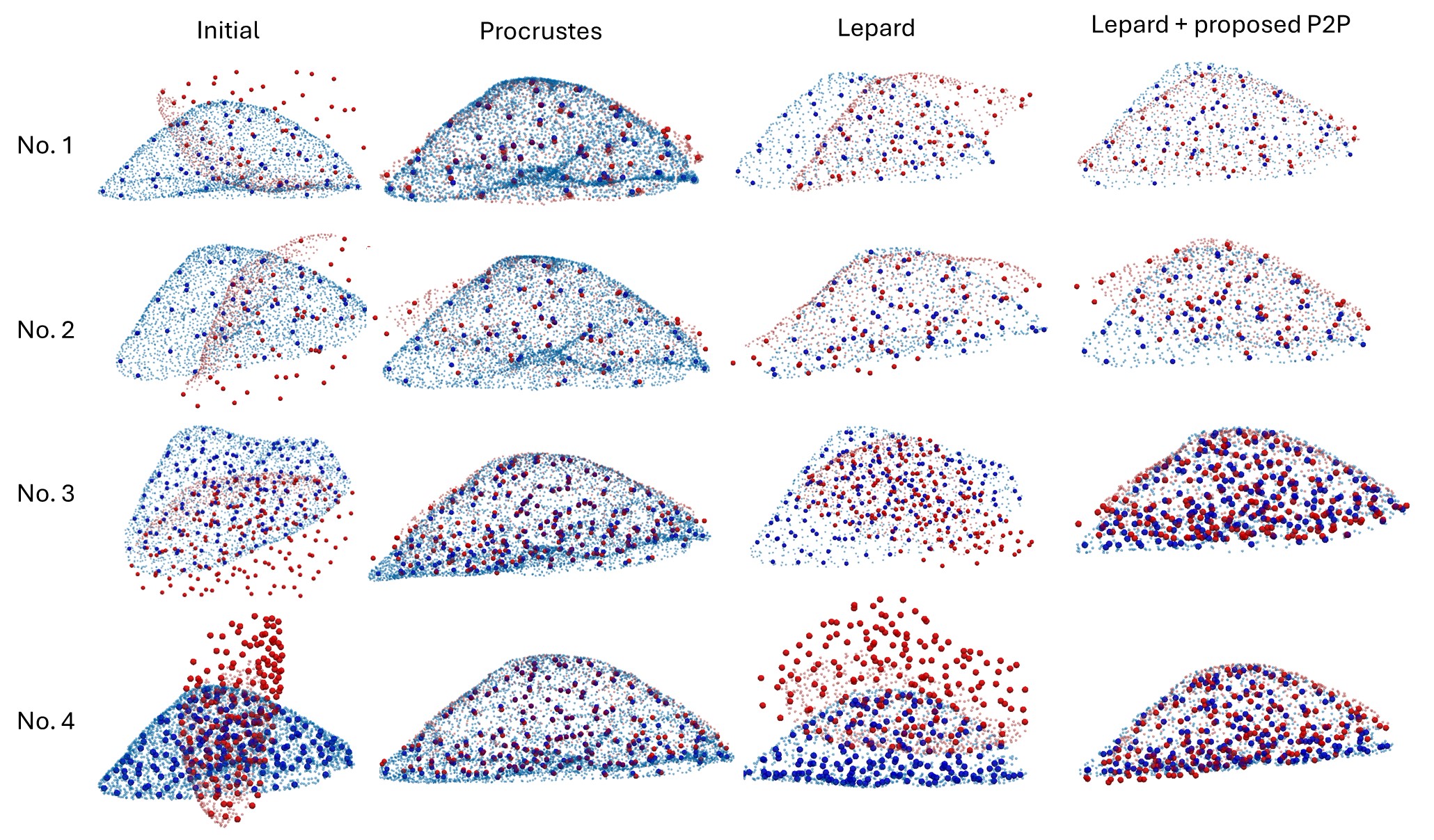}
    \caption{Qualitative comparison of registration results on the \textit{in vitro} phantom dataset. The source point cloud and its \textcolor{black}{volumetric} fiducial markers are shown in blue, while the target point cloud and its \textcolor{black}{volumetric} fiducial markers are shown in red, with fiducial markers displayed in larger sizes for visibility. }
    \label{fig:qualitative_phantom} 
\end{figure*}

% \begin{figure}[htb!]
%     \centering
%     \includegraphics[width=0.75\linewidth]{compare_phantom.png}
%     \caption{Comparison of registration errors from the \textit{In Vitro} phantom dataset, with the average RMS-TRE values plotted against different visibility ratios using the bin centers from Table \ref{tab:vis_phantom}.
%     }
%     \label{fig:Compare_Vitro_curves}
% \end{figure}

% and Fig. \ref{fig:Compare_sim_curves}

\subsubsection{Performance against Baselines}
We further compare the baseline methods with and without our proposed module using the phantom datasets. The results are reported in Table~\ref{tab:vis_phantom}. As seen, the \textit{In Vitro} results align with the findings from the \textit{In Silico} experiments. GO-ICP, RegTr, and RoITr are more sensitive to visibility variations compared to LiverMatch and Lepard. After applying our module, the RMS-TRE values for LiverMatch and Lepard are reduced from $20.54$ mm to $14.97$ mm and from $20.71$ mm to $12.45$ mm, respectively. These improvements are statistically significant.

\subsubsection{Visual Assessment}

For both the \textit{in silico} and \textit{in vitro} datasets, the target point clouds are automatically cropped. Additionally, we present qualitative results using manually cropped surfaces from \cite{yang2024boundary}, as shown in Fig.~\ref{fig:qualitative_phantom}. In these manually cropped cases, the proposed method still significantly improves the registration results compared to Lepard. 
% \textcolor{blue}{
Overall, we demonstrate that the versatility of our proposed module enables seamless plug-and-play integration to improve the robustness and reliability of deep learning methods for rigid point cloud registration.
% }

\subsection{Sensitivity and Ablation study}\label{subsec:ablation}

We further conduct comprehensive sensitivity and ablation studies on the \textit{in vitro} phantom dataset, incorporating the proposed module with Lepard to evaluate its effectiveness.

\textbf{Candidate patch number}. 
The only hyperparameter in the proposed module is the candidate patch number, $K$. We vary this hyperparameter and present the registration results in Fig. \ref{fig:ab_K}, which shows a clear trend of decreasing registration error as the number of candidate patches increases. Statistical significance was observed between $K=4$ and $K=5$, but not between $K=5$ and $K=6$, suggesting that the default value of $K=5$ provides an optimal balance between computational complexity and accuracy.

\textbf{Selection role}. We further compare our closest distance-based selection introduced in Eq. \ref{eq:closest}, with the inlier number-based selection in Eq. \ref{eq:inlier-based}, where the parameter $\tau$ is set to 0.05. As shown in Table \ref{tab:ablation}, our approach proves to be more effective, resulting in fewer errors below 2 mm.

\textbf{Patch candidate proposal}. We also investigate the contribution of the candidate proposal to our proposed module. As shown in Table \ref{tab:ablation}, omitting the patch candidate proposal results in a significant decline in performance, with a drop of approximately 2.5 mm in registration accuracy.

\textbf{Comparison with RANSAC}. Following previous studies \cite{yew2022regtr, yu2023rotation}, we further compare the proposed module with RANSAC, a typical robust estimator used in point cloud registration. As discussed in Section \ref{subsec:learning-based}, the robust estimators do not fit the end-to-end learning scheme and are not designed to handle the complete-to-partial ambiguity.

Using the correspondence-based RANSAC ICP implementation from Open3D \cite{Zhou2018}, with the hyperparameter for maximum correspondence distance set to 0.05, we found that our proposed module outperforms RANSAC in low-visibility scenarios, \textcolor{black}{achieving a RMS-TRE that is 3.10 mm lower than RANSAC’s result.}

RANSAC slightly worsened registration results in the visibility range $[0.4, 1)$, likely due to its random sampling nature.

\textcolor{black}{\textbf{Running time}.} The average running time of the proposed module \textcolor{black}{under the default setting ($K=5$) on the \textit{in vitro} dataset with the visibility [0.2, 0.3)} is 0.04 seconds, compared to 1.06 seconds for RANSAC.

\textcolor{black}{\textbf{Computational complexity}. The approximated computational complexity of our P2P strategy is \(\mathcal{O}(K \cdot M \cdot M)\), whereas a complete-to-partial registration has \(\mathcal{O}(N \cdot M)\).} \textcolor{black}{The complexity of P2P increases with \( K \), and both approaches have similar computational complexity when \( K \cdot M = N \).}

% Robust estimators can also improve registration results 

% Although the proposed module is not a robust estimator 

% \textcolor{black}{As discussed in Section \ref{subsec:learning-based}, RANSAC appears to be the most famous xxx method.}  

% We thus replaced the proposed module with RANSAC, and the comparison results are shown in Fig. \ref{fig:RANSAC_ours}. 

% \textcolor{black}{
% Also, as discussed in \S\ref{subsec:learning-based}, RANSAC often experiences slow convergence. Our findings indicate that the average running time of the proposed module is 0.04 seconds, compared to 1.06 seconds for RANSAC. This noticeable efficiency (i.e., 26.5$\times$ faster) of our method is crucial for real-time deployments.}

% Also, as we mentioned in \S\ref{subsec:learning-based}, RANSAC typically suffers from slow convergence. We found that the average running time of the proposed module is 0.04 seconds, compared to 1.06 seconds for RANSAC. This shows the efficiency of our proposed method, this is extremely important for real-time deployments. }

% \begin{table}[htb!]%[!htpb]
%     % \setlength{\tabcolsep}{4pt}
%     % \renewcommand{\arraystretch}{1.2}
% 	\centering
% 	\caption{Averaged registrations per second on the simulation dataset.}
% 	\label{tab:ablation}
% \resizebox{0.95\linewidth}{!}{	
% \begin{tabular}{l|c|c|c|c}
%  \hline
%  & w/ weighted SVD & w/ RANSAC ICP
% & w/ PCR
% \\
% \hline
% Lepard & 10.63 & 5.78 & 8.28
% \\
% LiverMatch & 14.07 & 1.85 & 11.48
% \\
% \hline
% \end{tabular}
% }	
% \end{table}

% \subsection{\textit{In Vivo}  Validation}
% \label{sec:vivo}

\section{Discussions}

\subsection{\textcolor{black}{Limitations of Baselines and Merits of P2P}}
\label{sec:dicsA}

While learning-based point cloud registration methods are grounded in a solid foundation in natural benchmarks, they face a unique challenge from the complete-to-partial ambiguity when applied to liver surgery scenarios, 
% \textcolor{blue}{
where only a partial surface is visible during an intraoperative operation.
% } 

To alleviate this ambiguity, we propose a P2P feature matching module to bridge the gap between methods and their applications in liver surgery. Also, as a comprehensive benchmark is missing in this field, we construct a benchmark consisting of \textit{in silico} and \textit{in vitro} datasets. 

% The constructed \textit{in silico} dataset contains a large number of liver models, and deformations may not only contribute to the learning-based rigid registration methods but also contribute to the learning-based non-rigid registration method.

We then evaluate representative correspondence-based methods on the constructed datasets. While these methods achieve comparable registration errors in high-visibility cases, their performance deteriorates as the visibility ratio decreases. 

\subsubsection{\textcolor{black}{Limitation of End-to-End Correspondence Design}} RegTr is particularly sensitive to reduced visibility, likely due to its unique design. RegTr directly predicts the displacements of source and target key points as correspondences, rather than matching features between the two point clouds. 
This approach works well when the source and target point clouds are of similar scale, but in cases of complete-to-partial ambiguity, predicting key point displacements is less robust due to the lack of global context from the target point cloud.

\subsubsection{\textcolor{black}{Limitation of Node-to-Group Strategy}} RoITr, a method based on the node-to-group strategy \cite{yu2021cofinet,qin2022geometric}, also shows sensitivity to low-visibility cases.  \textcolor{black}{This strategy does not consider the complete-to-partial ambiguity and extracts patches from both the source and target point clouds, with a hyperparameter controlling patch size. In contrast, our method represents a source point cloud with patches that contain the same number of points as the target. Also, its Superpoint Matching module to generate patch nodes at a coarse scale introduces misleading nodes due to coarse-scale errors and complete-to-partial ambiguity. In our approach, we first localize a region and then sample nodes from it, leading to more reliable patch selection. Moreover, we utilize a distance-based selection rule (Eq.~\ref{eq:closest}) to select the final rigid transformation, which demonstrates greater robustness than the inlier-based selection rule (Eq.~\ref{eq:inlier-based}) in handling complete-to-partial ambiguity (see Table \ref{tab:ablation}).}

\subsubsection{\textcolor{black}{Limitation of Complete-to-Partial Matching}} In comparison with RegTr and RoITr, LiverMatch and Lepard are less sensitive to decreasing visibility ratios. Both methods employ a U-Net-like structure and match features between the source and target point clouds at the final output, indicating that this simple pipeline may be more versatile. Despite showing promising results from LiverMatch and Lepard, their performances are still not satisfying in low-visibility cases.

\subsubsection{\textcolor{black}{Merits of the P2P Module}} \textcolor{black}{Given the above discussion, the designs of RegTr and RoITr are not well-suited for low-visibility cases. To address the complete-to-partial ambiguity, we developed the P2P module and integrated it into LiverMatch and Lepard.} 

In such instances, after applying our module, we observe significant improvements $w.r.t.$ the registration performance (see \textcolor{black}{key quantitative improvements} in Table \ref{tab:vis_sim}, \ref{tab:vis_phantom}). \textcolor{black}{For example, under low visibility range [0.2, 0.3), the approach improved registration errors for LiverMatch from 12.9 mm to 8.7 mm (-33\%) for its in silico datasets and from 20.5 mm to 15.0 mm (-27\%) for its in vitro datasets. Similarly, for Lepard, registration errors were improved from  9.5 mm to 6.7 mm (-29\%) for its in silico datasets and from 20.7 mm to 12.5 mm (-40\%) for its in vitro datasets.}

\textcolor{black}{Similar trends are seen under noise and deformation (Table \ref{tab:noise}, \ref{tab:deform_tre}). The P2P module consistently reduced RMS-TREs, with greater gains at higher noise. Under both low and high deformation ranges, P2P-enhanced methods showed lower errors and reduced variability. For instance, LiverMatch improved by $4.0$ mm and $4.7$ mm, and Lepard by $2.7$ mm and $2.7$ mm, respectively, for their in silico and in vitro datasets.}

\textcolor{black}{In addition to the significant improvement in terms of registration error, the P2P} has merits in:

\textbf{i)} Our proposed module alleviates complete-to-partial ambiguity by converting it to P2P matching. Our module only has one hyperparameter, $K$, whose physical meaning controls the number of proposed patches. From the parameter-sensitivity study shown in Fig. \ref{fig:ab_K}, increasing $K$ will improve the registration by expanding a fully parallelizable search space, and the default value $K=5$ yields satisfactory results. \textcolor{black}{The choice of \( K=5 \) is theoretically justified by the lowest characteristic visibility ratio, which is approximately 20\% \cite{heiselman2018characterization}. Given this minimum visibility, partitioning the complete point cloud into five patches ensures that at least one patch is likely to contain a corresponding structure in the partial target, even under the most challenging visibility conditions.}

In addition, we conducted an ablation study on our candidate proposal, and the closest point distance-based selection rule effectively contributes to the whole strategy. 

\textbf{ii)} Our proposed module can be seamlessly integrated with several learning-based correspondence methods to handle complete-to-partial ambiguity. The proposed module is differentiable, fast, and parameter-free, as it only involves feature resampling and matching and runs in parallel on the GPU. The proposed module contributes to advancing the robustness and reliability of fully automatic end-to-end point cloud registration.

% \textcolor{blue}{Robust estimators, like R}

% In contrast, RANSAC takes a much longer running time, may be trapped in a local minimum and is not differentiable.
% % , the comparison of which is included in the ablation study. 
% Moreover, it only filters outliers and does not generate new candidate matches, which is not specifically designed 
% for complete-to-partial ambiguity. 
% The proposed module, on the other hand, contributes to end-to-end point cloud registration without robust estimators.

% \textit{in vivo} validation will be required. I

\subsection{\textcolor{black}{Potential Challenges for Correspondence-based
Point Cloud Registration}}
\label{sec:dicsB}

\textcolor{black}{Correspondence-based
point cloud registration methods typically require that the source and target point clouds share similar properties, such as density and reconstruction accuracy. However, these prerequisites may be challenged in the context of liver surgery due to intraoperative constraints and variability in data acquisition.}

\textcolor{black}{One prerequisite is that the target point cloud represents a continuous liver surface region, allowing it to be voxelized to achieve a density similar to that of the source. However, this assumption may not hold when the target point cloud is manually collected using optical trackers or when other organs and surgical instruments partially occlude the intraoperative surface. To mitigate this issue, interpolation-based approaches, such as the grid-fitting method proposed by Collins \textit{et al.}\cite{collins2017improving}, or point cloud completion methods \cite{poudel2025evaluation} may help reconstruct a dense and continuous surface representation.}

\textcolor{black}{Minimizing reconstruction errors in both the source and target point clouds is crucial for accurate 3D-3D registration. In the source point cloud, errors may arise from liver segmentation from preoperative scans and the inherent limitations of scan resolution. For the target point cloud, inaccuracies often stem from depth sensor limitations and constraints in 3D surface reconstruction techniques. Despite these challenges, continuous advancements in hardware (e.g., high-precision depth sensors, CT) and software (e.g., stereo depth estimation algorithms\cite{disparity}) are steadily improving the feasibility of this approach for liver surgery.}

\subsection{\textcolor{black}{Limitations of Constructed Datasets}}
\label{sec:dicsC}

\textcolor{black}{
In our current \textit{in silico} dataset generation pipeline, the target point cloud is created by randomly cropping the full liver surface. However, in open and laparoscopic liver surgery, only the anterior liver surface is visible due to occlusions from other organs and surgical instruments. This discrepancy between synthetic and real-world visibility may hinder model generalizability. A cropping method that better simulates how a camera perceives the anterior liver surface during surgery could improve model performance when trained on the \textit{in vivo} dataset and testing on real surgical data.}  

\textcolor{black}{Our \textit{in silico} and \textit{in vitro} datasets primarily feature moderate deformations, similar to those in the Sparse Non-Rigid Registration Challenge dataset~\cite{heiselman2024image}. The deformation simulation pipeline~\cite{pfeiffer2020non} used to generate the \textit{in silico} dataset randomly applies zero boundary conditions and external forces, occasionally producing cases dominated by rigid transformations. Following the rigid alignment step described in \S\ref{subsubsec:in_silico_dataset}, we observed that the resulting deformations, quantified by RMS-TREs and visualized in Fig.~\ref{fig:vis_def}, are unevenly distributed and predominantly moderate.}

In future work, we aim to build a \textit{in vivo} dataset to further validate the performance of correspondence-based registration methods in real surgical scenarios, bringing our research closer to clinical applicability. \textcolor{black}{Insights gained from constructing the \textit{in vivo} dataset, such as the characteristic deformation levels observed in laparoscopic surgery and the zero-displacement boundary conditions imposed by anatomical structures, will inform refinements to the \textit{in silico} dataset, to improve training efficacy and generalization performance.}

\subsection{\textcolor{black}{Limitations of the P2P module}}
\label{sec:dicsD}

\textcolor{black}{To reduce redundant patches, the default setting of the P2P module generates patch nodes from the top candidate set, a design proven to be more effective than their selection from the entire set (see Table~\ref{tab:ablation}). However, this approach may produce candidate patches that fail to include the correct anatomical region in extreme cases, which may result from the prerequisites outlined in \S\ref{sec:dicsD} being unmet or due to extreme deformation. This issue may arise when the source and target features become highly inconsistent, leading to the failure of Visible Source Point Estimation in \S\ref{subsec:patches}. In such cases, disabling our patch candidate proposals and instead applying FPS directly to the anterior surface to generate patch nodes while increasing their number is advisable.
}

\textcolor{black}{Currently, the P2P module acts as an additional module to the networks. However, designing a network with a built-in P2P strategy may enhance efficiency and effectiveness. Exploring this approach is part of our future work.}

\subsection{\textcolor{black}{Clinical Significance}}
\label{sec:dicsD}

\textcolor{black}{The clinical significance of this work lies in translating learning-based point cloud registration techniques to streamline image-guided liver surgery workflows. Unlike conventional approaches that require manual interaction for initial rigid registration, learning-based methods have the potential to provide fully automatic and highly accurate alignment. This approach not only reduces operative time, but also facilitates initialization for subsequent non-rigid registration methods \cite{heiselman2020intraoperative,yang2024boundary, heiselman2024image} and aids surgeons in precisely localizing subsurface structures for more accurate tumor resection. Furthermore, given the anatomical similarities between the liver and other organs, such as the prostate and kidney, the proposed techniques and module could extend their benefits to image-guided surgeries for these organs as well.}

\section{Conclusion}

We have constructed a benchmark using both \textit{in silico} and \textit{in vitro} datasets to evaluate state-of-the-art learning-based point cloud matching methods. Our analysis underscores the challenge of complete-to-partial ambiguity in scenarios with low visibility ratios, as typically occurs during liver surgery. In light of this view, we introduce a P2P strategy as a module that can be seamlessly integrated with learning-based point cloud correspondence methods. We have demonstrated the effectiveness of this module in improving several methods, especially in low-visibility conditions. 

\textcolor{black}{We further discussed the limitations of state-of-the-art methods, our constructed datasets, and the proposed module in the context of \textit{in vivo} applications. To address these limitations, future work will focus on optimizing the cropping and deformation simulation in the \textit{in silico} dataset, constructing dedicated \textit{in vivo} datasets, exploring interpolation and point cloud completion techniques for improved intraoperative surface reconstruction, and developing a network that natively integrates the P2P strategy within its feature extraction pipeline.
}

To summarize, the constructed benchmark, the proposed module, and our comprehensive discussion lay a solid foundation for future studies of \textit{in vivo} applications.

% We have constructed a benchmark consisting of \textit{in silico} and \textit{in vivo} datasets, and comprehensively evaluated state-of-the-art learning-based correspondence point cloud matching methods. Our analysis highlights the challenge of complete-to-partial point cloud matching in liver surgery, primarily due to low visibility ratios. In light of this view, we introduced a patches-to-partial strategy in the form of a module that can be seamlessly plug-and-play with learning-based correspondence methods. The proposed module has proven effective in enhancing several methods in low-visibility cases.

% Overall, our work offers a valuable resource and a practical solution that advances the accuracy of point cloud matching in challenging surgical applications.

% This dataset, along with our findings and proposed strategy, lays a solid foundation for future studies in \textit{in vivo} applications.

\bibliographystyle{ieeetr}
\bibliography{refs}

\end{document}